\documentclass[11pt]{article}

\usepackage{lineno,hyperref}
\modulolinenumbers[5]

\usepackage{amsfonts}
\usepackage{amsmath}
\usepackage{listings}
\usepackage{algorithm}
\usepackage{algorithmic}
\usepackage{booktabs}
\usepackage{gensymb}
\usepackage{subcaption}
\usepackage[toc,page]{appendix}
\usepackage{xcolor}
\usepackage{graphicx}
\usepackage{arxiv}
\usepackage{setspace}

\newcommand{\ra}[1]{\renewcommand{\arraystretch}{#1}}

\DeclareMathOperator*{\argmin}{arg\,min}

\title{Gym-ANM: Reinforcement Learning Environments for Active Network Management Tasks in Electricity Distribution Systems}

\date{}
\author{
  Robin~Henry \\
  The University of Edinburgh\\
  \texttt{robin@robinxhenry.com} \\
   \And
 Damien~Ernst \\
  The University of Liège \\
  \texttt{dernst@uliege.be} \\
}

\begin{document}
\maketitle

\begin{abstract}
Active network management (ANM) of electricity distribution networks include many complex stochastic sequential optimization problems. These problems need to be solved for integrating renewable energies and distributed storage into future electrical grids. In this work, we introduce Gym-ANM, a framework for designing reinforcement learning (RL) environments that model ANM tasks in electricity distribution networks. These environments provide new playgrounds for RL research in the management of electricity networks that do not require an extensive knowledge of the underlying dynamics of such systems. Along with this work, we are releasing an implementation of an introductory toy-environment, ANM6-Easy, designed to emphasize common challenges in ANM. We also show that state-of-the-art RL algorithms can already achieve good performance on ANM6-Easy when compared against a model predictive control (MPC) approach. Finally, we provide guidelines to create new Gym-ANM environments differing in terms of (a) the distribution network topology and parameters, (b) the observation space, (c) the modelling of the stochastic processes present in the system, and (d) a set of hyperparameters influencing the reward signal. Gym-ANM can be downloaded at \url{https://github.com/robinhenry/gym-anm}.
\end{abstract}


\section{Introduction}
Reinforcement learning (RL) is a vibrant field of machine learning aiming to mimic the human learning process. This allows us to solve numerous complex decision-making problems \cite{sutton2018reinforcement}. In the field of power systems (a term used to refer to the management of electricity networks), researchers and engineers have used RL techniques for many years \cite{glavic2017reinforcement}. Over the last few years, however, decision-making challenges in power systems have drawn less attention than other domains in which RL has been successfully and extensively applied, such as the fields of games \cite{mnih2013playing, mnih2015human, silver2016mastering, vinyals2019grandmaster}, robotics \cite{deisenroth2013survey, kormushev2013reinforcement, kober2013reinforcement, gu2017deep}, and autonomous driving \cite{sallab2017deep, o2018scalable, li2019reinforcement}. A plausible explanation for this is the lack of off-the-shelf simulators that model such problems. Indeed, despite its many recent breakthroughs, RL research remains largely dependent on the availability of artificial simulators that can be used as surrogates for the real world \cite{dulac2019challenges}. Training on real systems is often too slow and constraining, while simulators allow us to take advantage of large computational resources and do not constrain exploration.

Developing efficient and reliable algorithms to solve decision-making challenges in power systems is becoming more and more crucial for ensuring a smooth transition to sustainable energy systems. Power grids have experienced profound structural and operational changes over the last two decades \cite{fang2011smart}. The liberalization of electricity markets introduced a competitive aspect in their management, driving network improvements and cheaper energy generation \cite{joskow2008lessons}. The arrival of distributed generators, such as wind turbines and photovoltaic panels (PVs), has compromised the traditional model of decentralized generation. In particular, we have seen the appearance of (virtual) microgrids creating local energy ecosystems in which consumers are now also producers \cite{lasseter2002microgrids}. In the near future, we can expect the addition of even more distributed generators to the grid \cite{capitanescu2014assessing}, along with an increase in the number of large loads due to the fast electric vehicle market growth \cite{lutsey2016sustaining}. Power grids are also facing the emergence of distributed energy storage (DES), with certain technologies already available, such as batteries \cite{divya2009battery} and power-to-gas \cite{gotz2016renewable}. As a result, system operators are facing many new complex decision-making problems (overvoltages, transmission line congestion, voltage coordination, investment issues, etc.), some of which might benefit from advances in the very active area of RL research.

Through this work, we seek to promote the application of RL techniques to active network management (ANM) problems, a class of sequential decision-making tasks in the management of electricity distribution networks (DNs). In the power system literature, ANM refers to the design of control schemes that modulate the generators, the loads, and/or the DES devices connected to the grid. This is done to avoid problems at the distribution level and maximize profitability through, e.g., avoidable energy loss \cite{gill2013dynamic}. This modulation, operated by distribution network operators (DNOs), may result in a necessary reduction in the output of generators from what they could otherwise have produced given available resources, often referred to as the process of curtailment. Such generation curtailment, along with storage and transmission losses, constitute the principal sources of energy loss that we would like to minimize through ANM. At the same time, the ANM scheme must ensure a safe and reliable operation of the DN. This is often expressed as a set of operational constraints that must be satisfied.

More specifically, we propose Gym-ANM, a framework that facilitates the design and the implementation of RL environments that model ANM tasks. Our goal was to release a tool that could be used without an extensive background in power system analysis. We thus engineered Gym-ANM so as to abstract away most of the complex dynamics of power system modelling. With its different customizable components, Gym-ANM is a suitable framework to model a wide range of ANM tasks, from simple ones that can be used for educational purposes, to complex ones designed to conduct advanced research. In addition, Gym-ANM is built on top of the OpenAI Gym toolkit \cite{brockman2016openai}, an interface with which a large part of the RL community is already familiar. Note that Gym-ANM environments do not solve ANM problems but, rather, provide a simple programming interface to test and compare various optimization and RL algorithms that aim to do so.

The remainder of this paper is organized as follows. First, we introduce a series of background concepts and notations in RL, DNs, and MPC in Section \ref{sec:background}. Section \ref{sec:gym-anm} then formalizes the generic ANM task that we consider as a partially observable Markov decision process (POMDP). Next, we propose a specific Gym-ANM environment that highlights common ANM challenges, ANM6-Easy, in Section \ref{sec:environments}. The performance of the state-of-the-art proximal policy optimization \cite{schulman2017proximal} (PPO) and soft actor-critic \cite{haarnoja2018soft} (SAC) deep RL algorithms are evaluated on ANM6-Easy in Section \ref{sec:experiments}. Finally, Section \ref{sec:conclusion} concludes our work. To keep this paper accessible to a broad audience and to provide interested readers with a formal introduction to power system modelling, technical details about the inner working of the power grid simulator are gathered in Appendices \ref{app:simulator} and \ref{app:opf}. Guidelines to design and implement new Gym-ANM environments are also provided in Appendices \ref{app:new_env} and \ref{app:case_file}, and more in-depth tutorials and documentation can be found on the project repository at \url{https://github.com/robinhenry/gym-anm}.

\begin{figure}[]
    \centering
    \fbox{\includegraphics[width=\linewidth]{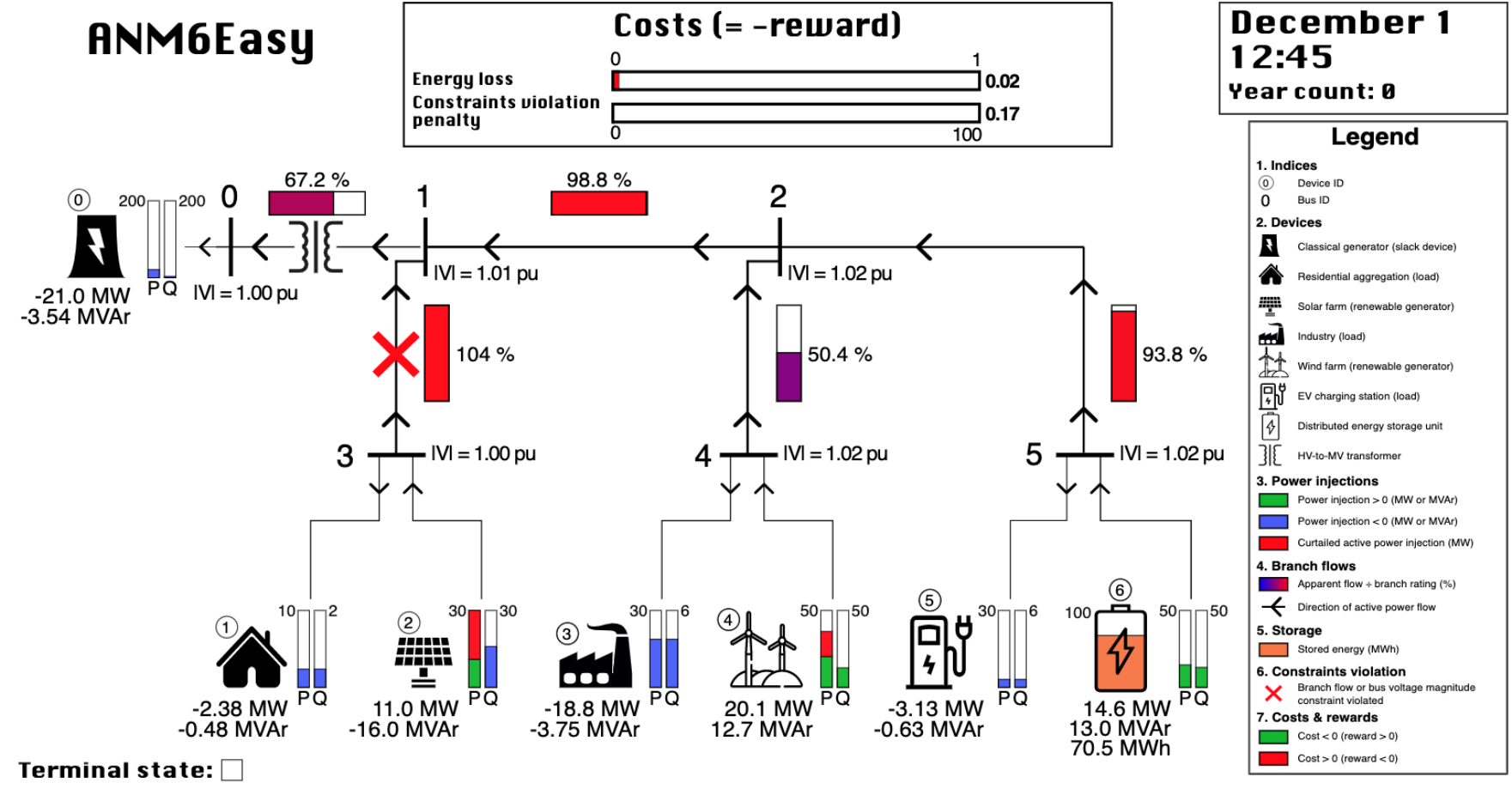}}
    \caption{A Gym-ANM environment. At this specific time, the agent is
    curtailing both renewable energy resources and discharging the DES unit. Transmission line 1-3 is overheating with a power flow of 104\% of its capacity.}
    \label{fig:env_example}
\end{figure}

\section{Background}
\label{sec:background}
\subsection{Reinforcement Learning}
\label{sec:background_rl}
We consider the standard RL setting for continuing tasks where an agent interacts with an environment $E$ over an infinite sequence of discrete timesteps $\mathcal T = \{0,1,\ldots\}$, modelled as a Markov decision process (MDP). At each timestep $t$, the agent selects an action $a_t \in \mathcal A$ based on a state $s_t \in \mathcal S$ according to a stochastic policy $\pi:\mathcal S \times \mathcal A \to [0, 1]$, such that $a_t \sim \pi(\cdot|s_t)$. After the action is applied, the agent transitions to a new state $s_{t+1} \sim p(\cdot|s_t, a_t) \in \mathcal S $ and receives the reward $r_t = r(s_t, a_t, s_{t+1}) \in \mathbb R$. The return from state $s_t$ is defined as $R_t = \lim_{T\to \infty} \sum_{i=t}^{T-1} \gamma^{i-t}r_t$, where $\gamma \in [0,1)$ is the discount factor determining the weight of short- versus long-term rewards. We also distinguish a set of terminal states $\mathcal S^{terminal} \subset \mathcal S$. Given the distribution of initial states $p_0(\cdot)$ and the set of stationary policies $\Pi$, a policy $\pi \in \Pi$ is considered optimal if it maximizes the expected return $J_{\pi}(s_0) = \mathbb E_{s_{i>0}, r_{i \ge 0} \sim E, a_i \sim \pi} [R_0|s_0]$ for all $s_0$ that belong to the support of $p_0(\cdot)$, and where rewards received after reaching a terminal state are always zero. In the context of problems with large and/or continuous state-action spaces, RL often focuses on learning a parameterized policy $\pi_\phi \in \Pi$ with parameters $\phi$ whose expected return $J_{\pi_\phi}(s_0)$ is as close as possible to that of an optimal policy.

In many cases, the environment may be partially observable so that the agent only has access to observations $o \in \mathcal O$. The agent must thus adequately infer, directly or indirectly, an approximation of the state $s_t$ from the history of observation-action-reward tuples $h_t = (o_0,a_0,r_0\ldots, a_{t-1},r_{t-1}, o_t)$. We designed the Gym-ANM framework so that it is straightforward for researchers to experiment with different degrees of observability in each environment.

\subsection{Distribution Networks}
\label{sec:background_dn}
An electricity distribution network can be represented as a directed graph $G(\mathcal N, \mathcal E)$, where $\mathcal N=\{0,1, \dots, N-1\}$ is a set of positive integers representing the buses (or nodes) in the network, and $\mathcal E \subseteq \mathcal N \times \mathcal N$ is the set of directed edges linking buses together. The notation $e_{ij} \in \mathcal E$ refers to the directed edge with sending bus $i$ and receiving bus $j$. Each bus might be connected to several electrical devices, which may inject into or withdraw power from the grid. The set of all devices is denoted by $\mathcal D = \{0, 1, \ldots, D-1\}$, the set of all devices connected to bus $i \in \mathcal N$ by $\mathcal D_i \subseteq \mathcal D$, and it is assumed that each device is connected to a single bus. 

Several variables (complex phasors) are associated with each bus $i \in \mathcal N$: a bus voltage level $V_i$, a bus current injection $I_i$, an active (real) power injection $P_i^{(bus)}$, and a reactive power injection $Q_i^{(bus)}$. The bus power injections $P_i^{(bus)}$ and $Q_i^{(bus)}$ can also be obtained from $P_i^{(bus)} = \sum_{d \in \mathcal D_i} P_d^{(dev)}$ and $Q_i^{(bus)} = \sum_{d \in \mathcal D_i} Q_d^{(dev)}$, where $P^{(dev)}_d$ and $Q^{(dev)}_d$ denote the active and reactive power injections from device $d \in \mathcal D$ into the grid, respectively. The complex powers $S_i^{(bus)}, S_d^{(dev)} \in \mathbb C$ injected into the network at bus $i$, or device $d$, can then be obtained from the relation $S_i^{(bus)} = P_i^{(bus)} + \mathbf i Q_i^{(bus)}$ or $S_d^{(dev)} = P_d^{(dev)} + \mathbf i Q_d^{(dev)}$. Similarly, variables $I_{ij}, P_{ij}, Q_{ij},$ and $S_{ij}$ refer to the directed flow of these quantities in branch $e_{ij} \in \mathcal E$, as measured at bus $i$. Note that, as a result of transmission losses, power and current flows may have different magnitudes at each end of the branch, e.g. $|P_{ij}| \neq |P_{ji}|$.

\subsection{Model Predictive Control (MPC) and Optimal Power Flow (OPF)}
In this work, we also present a model predictive control (MPC) approach to solving the ANM tasks that we propose with Gym-ANM. MPC in discrete-time settings is a control strategy in which, based on a known model of the dynamics of the system, a multi-stage optimization problem is solved at each timestep over a finite time horizon. The solution found is applied to the system at the current timestep, and the process is repeated at the next one, indefinitely \cite{camacho2013model}. The fact that a multi-stage optimization problem based on a model of the system is solved at each time step allows MPC to plan ahead and anticipate the system's behavior. This leads to near-optimal performance as the optimization horizon is increased (assuming an accurate model of the system).

The optimization problem solved by our MPC control algorithm is a multi-stage optimal power flow (OPF) problem. Since its first formulation by Carpentier in 1962 \cite{carpentier1962contribution}, solving a single instance or multiple instances of the OPF problem at regular time intervals has been the dominant approach to tackling decision-making problems in the management of power systems when network constraints are taken into account. In its most general form, the OPF problem is a non-convex constrained optimization problem with equality and inequality constraints. The objective function to minimize is often a representation of network operating costs, the equality constraints model the physical flows of electricity, and the inequality constraints model operational constraints. There exist many different formulations of the OPF problem, each designed to solve a particular control task in power systems. Although many solution methods have been proposed using a wide range of optimization tools and techniques, no single formulation has been accepted as suitable for all forms of OPF problems and it remains an active area of research. For the interested reader, comprehensive surveys of such approaches can be found in \cite{frank2012optimal1, frank2012optimal2}.

\section{Gym-ANM}
\label{sec:gym-anm}
In this section, we propose Gym-ANM, a framework that can model a wide range of novel sequential decision-making ANM tasks to be solved by RL agents. Each Gym-ANM task is provided as a Gym \cite{brockman2016openai} environment $E$ that we describe by the MDP $(\mathcal S, \mathcal A, \mathcal O, p_0, p, r, \gamma)_E$. Our formalization of these MDPs follows closely, and was inspired by, the work of Gemine et al. in \cite{gemine2017active}.

For mathematical convenience, the set of electrical devices $\mathcal D$ connected to the grid is divided into three disjoint subsets $\mathcal D_G$, $\mathcal D_L$, and $\mathcal D_{DES}$, so that $|\mathcal D_G| + |\mathcal D_L| + |\mathcal D_{DES}| = |\mathcal D|$. The set $\mathcal D_G$ contains the generators, $\mathcal D_L$ the loads, and $\mathcal D_{DES}$ the DES units. Generators represent devices that only inject power into the grid, such as renewable energy resources (RER) $\mathcal D_{RER} \subset \mathcal D_{G}$ or other traditional power plants $\mathcal D_G - \mathcal D_{RER}$. Loads group the passive devices that only withdraw power from the grid. Storage units, on the other hand, can both inject and withdraw power into/from the network. The only exception is the slack generator $g^{slack} \in \mathcal D_G - D_{RER}$, assumed to be the only device connected to the slack bus. The slack bus is a special bus used to balance power flows in the network and provide a voltage reference. The slack bus can also either inject or withdraw power into/from the network, such that the total generation remains equal to the total load plus transmission losses, at all times.

\subsection{Overview}
\label{sec:gym_anm_overview}
\begin{figure}[h]
  \includegraphics[width=\linewidth]{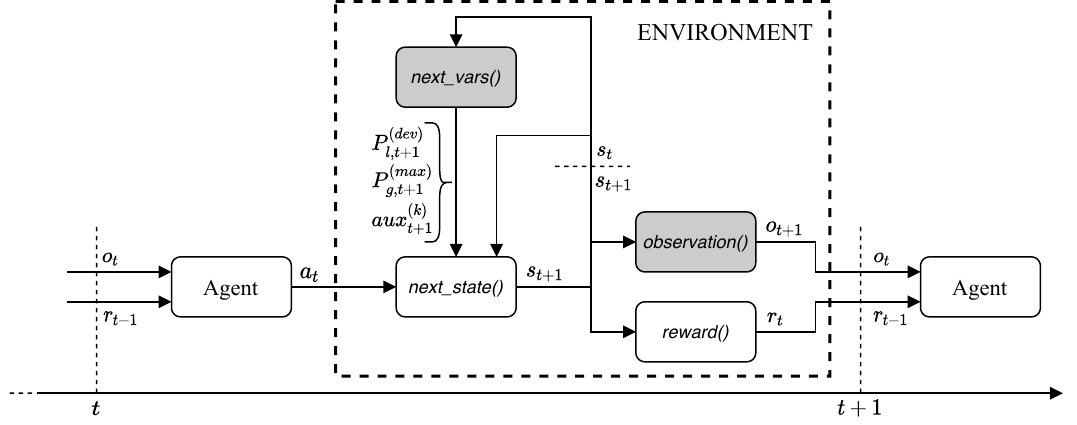}
  \caption{The Gym-ANM framework.}
  \label{fig:env_structure}
\end{figure}

The structure of the Gym-ANM framework is illustrated in Figure \ref{fig:env_structure}, in which grey blocks represent components (functions) that are fully customizable by the user to design unique ANM tasks. At each timestep $t$, the agent receives an observation $o_t \in \mathcal O$ and a reward $r_{t-1} \in \mathbb R$, based on which it then selects an action $a_t \in \mathcal A$ to be applied in the environment. 

Once the environment has received the selected action $a_t$, it samples a series of internal variables using the \texttt{next\_vars()} generative process conditioned on the current state $s_t \in \mathcal S$. These internal variables model the temporal stochastic evolution of the electricity demand and of the maximum renewable energy production before curtailment across the DN, as further described in later sections. 

The internal variables are then passed, along with $a_t$ and $s_t$, to the main function \texttt{next\_state()}, which applies the action to the environment and outputs the new state $s_{t+1} \in \mathcal S$. The \texttt{next\_state()} block behaves deterministically for a given DN. It first maps the selected action $a_t$ to the current available action space $\mathcal A(s_t) \subseteq \mathcal A$ before applying it to the environment. All the currents, voltages, energy storage levels, and power flows and injections are then updated, resulting in a new state $s_{t+1}$. Most of the power system modelling of the environment is handled by the \texttt{next\_state()} component, which we provide as a built-in part of the framework.

The new state $s_{t+1}$ is then used to compute the new observation $o_{t+1} \in \mathcal O$ and reward $r_t \in \mathbb R$. Much like the \texttt{next\_vars()} block, the behavior of the \texttt{observation()} component can be freely designed by the designer of the environment. This way, it becomes straightforward to investigate the impact of different observation vectors on the performance of a given algorithm on a given ANM task. To simplify the use of our framework, we also provide a set of default common observation spaces that researchers can experiment with. 

Our framework provides a built-in \texttt{reward()} component that computes the reward $r_t$ as:
\begin{align}
r_t = clip(-r_{clip},-(\Delta E_{t:t+1}+ \lambda \phi(s_{t+1})), r_{clip}),    
\end{align}
where $\Delta E_{t:t+1}$ is the total energy loss during $(t,t+1]$, $\phi(s_{t+1})$ is a penalty term associated with the violation of operating constraints, $\lambda$ is a weighting hyperparameter, and $r_{clip} > 0$ keeps the rewards within a finite range $[-r_{clip}, r_{clip}]$. This reward function was designed to reflect the overall goal: learn a control policy $\pi$ that ensures a secure operation of the DN while minimizing its operating costs. In the management of real-world DNs, there are many varied sources of operating costs. For simplicity, however, we consider energy losses and the violation of operational constraints to be the only sources of costs. Our reward formulation also assumes that the action is selected by the agent at time $t$, immediately applied in the environment at time $t+\epsilon$, with $\epsilon \to 0$, and that all power injections remain constant during $(t+\epsilon, t+1]$.

The Gym-ANM framework allows for the creation of environments that model highly customizable ANM tasks. In particular, varying any of the following components will result in a different MDP, and therefore a different ANM task: 
\begin{enumerate}
    \item{\bf Topology and characteristics of the DN.} Its topology is described by the tuple $(\mathcal D, \mathcal N, \mathcal E)$ and its characteristics refer to the parameters of each of its device $d \in \mathcal D$, bus $i \in \mathcal N$, and transmission link $e_{ij} \in \mathcal E$. In particular, the number of devices $|\mathcal D|$ and their respective operating range will shape the resulting state space $\mathcal S$ and action space $\mathcal A$. A detailed list of all the DN parameters modelled in Gym-ANM is provided in Appendix \ref{app:case_file}.
    \item{\bf Stochastic processes.} This corresponds to the design of the \texttt{next\_vars()} component in Figure \ref{fig:env_structure}. This component must model the temporal evolution of the electricity demand $P_{l,t}^{(dev)}$ of each load $l \in \mathcal D_L$, the maximum production $P_{g,t}^{(max)}$ that each generator $g \in \mathcal D_G - \{g^{slack}\}$ could produce at time $t$ (before curtailment is applied if $g \in \mathcal D_{RER}$), and a set of $K$ auxiliary variables $\{aux_t^{(k)}\}_{k=0}^{K-1}$.
    \item{\bf Observation space.} The observation space $\mathcal O$ can be changed to make the task more or less challenging for the agent by modifying the \texttt{observation()} function.
    \item{\bf Hyperparameters.} Although the \texttt{reward()} component is built-in as a part of the Gym-ANM framework, it nonetheless relies on three hyperparameters that can be chosen for each new task: the penalty weighting hyperparameter $\lambda$, the amount of time $\Delta t$ (in fraction of hour) elapsed between subsequent discretization timesteps, and the clipping hyperparameter $r_{clip}$. Because we consider a policy to be optimal if it minimizes the expected sum of discounted costs, we also consider the discount factor $\gamma \in [0, 1)$ to be another fixed hyperparameter part of the task description.
\end{enumerate}

In the remainder of this section, we explore the resulting MDP in more detail.

\subsection{State Space}
\label{sec:state_space}
At any timestep $t$, the state of a Gym-ANM environment is fully described by the state of the DN that it models. We represent this state using a set of state variables aggregated into a vector $s_t \in \mathcal S$:
\begin{align}
    s_t = \big[\{P_{d,t}^{(dev)}\}_{d \in \mathcal D},\; \{Q_{d,t}^{(dev)}\}_{d \in \mathcal D},\; \{SoC_{d,t}\}_{d \in \mathcal D_{DES}}, \{P_{g,t}^{(max)}\}_{g \in \mathcal D_G - \{g^{slack}\}},\; \{aux^{(k)}_t\}_{k =0}^{K-1} \big] \;, \label{eq:state}
\end{align}
where 
\begin{itemize}
    \item $P_{d,t}^{(dev)}$ and $Q_{d,t}^{(dev)}$ refer to the active and reactive power injections of device $d \in \mathcal D$ into the grid, respectively,
    \item $SoC_{d,t}$ is the charge level, or state of charge (SoC), of DES unit $d \in \mathcal D_{DES}$,
    \item $P_{g,t}^{(max)}$ is the maximum production that generator $g \in \mathcal D_G - \{g^{slack}\}$ can produce,
    \item $aux^{(k)}_t$ is the value of the $(k-1)$\textsuperscript{th} auxiliary variable generated by the \texttt{next\_vars()} block during the transition from timestep $t$ to timestep $t+1$.
\end{itemize}

In (\ref{eq:state}), the first $2|\mathcal D| + |\mathcal D_{DES}|$ variables $(P_{0,t}^{(dev)}, \ldots, SoC_{|\mathcal D_{DES}|-1, t})$ can be used to compute any other electrical quantities of interest in the DN (i.e., currents, voltages, power flows and injections, and energy storage levels), as derived in Appendix \ref{app:simulator}. We also include the maximum generation variables $P_{g,t}^{(max)}$ in $s_t$ because, even though they do not affect the physical electric flows in the network, they are required to compute the reward signal (see Section \ref{sec:reward_function}). 

These variables do not, however, provide any information about the temporal behavior of the system. Hence, they are not sufficient to describe the full state of the system from a Markovian perspective. For instance, it may not be enough to know the active and reactive power injections from a load $l \in \mathcal D_L$ at time $t$ to fully describe the probability distribution of its next demand $P_{l,t+1}^{(dev)}$.

In order to make $s_t$ Markovian, we chose to include a set of $K$ auxiliary variables $\{aux_t^{(k)}\}_{k=0}^{K-1}$ that can be used to model other temporal factors that influence the outcomes $P_{l,t+1}^{(dev)}$ and $P_{g,t+1}^{(max)}$ during the \texttt{next\_vars()} call of Figure \ref{fig:env_structure}. This leads to state transitions that are only conditioned on the current state of the environment and on the action the agent selects, i.e., $s_{t+1} \sim p(\cdot|s_t, a_t)$. The overall task is thus indeed a MDP.

For example, the environment ANM6-Easy that we introduce in Section \ref{sec:environments} uses a single auxiliary variable that represents the time of the day. This is sufficient to make $s_t$ Markovian, since the underlying stochastic processes can all be expressed as a function of the time of day. Another example would be an environment in which the next demand of each load and the generation from each generator is solely dependent on their current value. In this case, $s_t$ would not require any extra auxiliary variables. As environments become more and more complex, we expect state vectors to contain many auxiliary variables. Such examples could include solar irradiation and wind speed information to better represent the evolution of the electricity produced by renewable energy resources.

Finally, the environment may also reach a terminal state $s_t \in \mathcal S^{terminal}$, indicating that no solution to the power flow equations (see Appendix \ref{app:network_equations}) was found as a result of the action taken by the agent. This means that the power grid has collapsed and is often due to a voltage collapse problem \cite{chiang1990voltage}.

\subsection{Action Space}
\label{sec:action_space}
Given the current state of the environment $s_t \in \mathcal S$, the available actions are denoted by the action space $\mathcal A(s_t)$. We define an action vector $a_t \in \mathcal A(s_t)$ as:
\begin{align}
    a_t = \big[\{a_{P_{g,t}} \}_{g \in \mathcal D_G - \{g^{slack}\}},\; \{a_{Q_{g,t}} \}_{g \in \mathcal D_G - \{g^{slack}\}}, \{a_{P_{d,t}}\}_{d \in \mathcal D_{DES}},\; \{a_{Q_{d,t}}\}_{d \in \mathcal D_{DES}} \big] \;, \label{eq:action_vector}
\end{align}
for a total of $N_a = 2|\mathcal D_G| + 2|\mathcal D_{DES}| - 2$ control variables to be chosen by the agent at each timestep. Each control variable belongs to one of four categories:
\begin{itemize}
    \item{$a_{P_{g,t}}$:} an upper limit on the active power injection from generator $g \in \mathcal D_G - \{g^{slack}\}$. If $g \in \mathcal D_{DER}$, then $a_{P_{g,t}}$ is the curtailment value. For classical generators, it simply refers to a set-point chosen by the agent. The slack generator is excluded, since it is used to balance load and generation and, as a result, its power injection cannot be controlled by the agent. That is, $g^{slack}$ will inject the amount of power needed to fill the gap between the total generation and demand into the network.
    \item{$a_{Q_{g,t}}$}: the reactive power injection from each generator $g \in \mathcal D_G - \{g^{slack}\}$. Again, the injection from the slack generator is used to balance reactive power flows and cannot be controlled by the agent.
    \item{$a_{P_{d,t}}$}: the active power injection from each DES unit $d \in \mathcal D_{DES}$.
    \item{$a_{Q_{d,t}}$}: the reactive power injection from each DES unit $d \in \mathcal D_{DES}$.
\end{itemize}

The resulting action space $\mathcal A(s_t)$ is bounded by three sets of constraints. First, individual control variables in $a_t \in \mathcal A(s_t)$ are restricted to finite ranges $[\underline P, \overline P]$ or $[\underline Q, \overline Q]$. This is because electrical devices cannot physically inject (withdraw) infinite active or reactive power into (from) the network. Second, generators and DES units may have additional constraints on their current injections, such as current limits of power converters. These constraints further restrict the range of $(P, Q)$ injection points that these devices can apply, i.e. they cannot simultaneously operate at full capacity for both active and reactive power. Third, the range of possible active power injection from each DES unit depends on its current storage level (provided in $s_t$). Indeed, empty (full) units cannot inject (withdraw) any power into (from) the network. Note that the first two sets of constraints remain the same for all $s_t \in \mathcal S$ (see Appendix \ref{app:electrical_devices}).

For simplicity, the agent is never given the precise boundaries of the action space $\mathcal A(s_t)$. Instead, we let it choose an action within a larger set $\mathcal A$ bounded only by the first set of constraints, i.e. $\mathcal A$ ignores current limits in generators and DES units, as well as storage levels. In the case where the agent selects an action $a_t \in \mathcal A$ that falls outside of the current action space $\mathcal A(s_t)$, the action that is actually applied in the environment during the \texttt{next\_state()} call is the action in $\mathcal A(s_t)$ that stands the closest to $a_t$, according to the Euclidean distance (see Appendix \ref{app:transition_function}).

As a result, $\mathcal A$ is always bounded. Its bounds can be retrieved by the agent through the built-in \texttt{action\_space()} function. This allows users to follow good practices by working with agents that generate normalized action vectors in $[-1, 1]^{N_a}$.

\subsection{Observation Space}
\label{sec:obs_space}
In general, DNOs rarely have access to the full state of the distribution network when doing ANM. To model these real-world scenarios, Gym-ANM allows the design of a unique observation space $\mathcal O$ through the implementation of the \texttt{observation()} component, which may result in a partially observable task. We only assume that the size of $o_t$ remains constant.

To simplify the design of customized observation spaces, Gym-ANM also allows researchers to simply specify a set of variables to include in the observation vectors (e.g., branch active power flows $\{P_{12},P_{23}\}$ and bus voltage magnitudes $\{|V_0|, |V_2|\}$) of the new environment. The full list of available variables from which to choose is given in Appendix \ref{app:new_env}.

The agent can access the bounds of the observation space through the function call \texttt{observation\_space()}. This functionality may be of particular interest to agents that use neural networks to learn ANM policies, in which case normalized input vectors may increase training speed and stability.

\subsection{Transition Function}
\label{sec:transition_function}

Each state transition occurs in two steps. First, the outcomes of the internal stochastic variables $\{P_{l,t+1}^{(dev)}\}_{l \in \mathcal D_L}$, $\{P_{g,t+1}^{(max)}\}_{g \in \mathcal D_G - \{g^{slack}\}}$, and $\{aux_{t+1}^{(k)}\}_{k=0}^{K-1}$ are generated by the \texttt{next\_vars()} block of the Gym-ANM framework (see Figure \ref{fig:env_structure}). Once the selected action $a_t \in \mathcal A$ has been passed to the environment, the remainder of the transition is handled by the \texttt{next\_state()} component in a deterministic way. The reactive power injection of each load $d \in \mathcal D$ is directly inferred from its active power injection (assuming a constant power factor). The action $a_t$ is then mapped to $\mathcal A(s_t)$ according to the Euclidean distance and applied in the environment. Finally, all electrical quantities are updated by solving a set of so-called network equations (see Appendix \ref{app:network_equations}). The computational steps taken by \texttt{next\_state()} are described in more detail in Appendix \ref{app:transition_function}.

\subsection{Reward Function}
\label{sec:reward_function}

The reward signal is implemented by the built-in \texttt{reward()} block of Figure \ref{fig:env_structure} and is given by:
\begin{align}
    r_t =
    \begin{cases}
        clip(-r_{clip}, c_t, r_{clip}), & \text{if } s_{t+1} \notin \mathcal S^{terminal}, \\
        - \frac{r_{clip}}{1 - \gamma}, & \text{if } s_t \notin \mathcal S^{terminal} \text{ and }  s_{t+1} \in \mathcal S^{terminal}, \\
        0, & \text{else,}
    \end{cases}
    \label{eq:full-reward}
\end{align}
where
\begin{align}
    c_t = -(\Delta E_{t:t+1} + \lambda \phi(s_{t+1})).
    \label{eq:reward-ct}
\end{align}
Using a reward clipping parameter $r_{clip}$ ensures that any transition from a non-terminal state to a terminal one (i.e., when the power grid collapses), generates a much larger reward than any other transition does. As a result, it encourages the agent to learn a policy that avoids such scenarios at all costs. Subsequent rewards are always zero, until a new trajectory is started by sampling a new initial state $s_0$. 

During all other transitions, the energy loss $\Delta E_{t:t+1}$ is computed in three parts:
\begin{align}
\Delta E_{t:t+1} = \Delta E_{t:t+1}^{(1)} + \Delta E_{t:t+1}^{(2)} + \Delta E_{t:t+1}^{(3)} \;,
\end{align}
where:
\begin{itemize}
    \item $\Delta E_{t:t+1}^{(1)}$ is the total transmission energy loss during $(t, t+1]$. This is a result of leakage in transmission lines and transformers.
    \item $\Delta E_{t:t+1}^{(2)}$ is the total net amount of energy flowing from the grid into DES units during $(t, t+1]$. Over a sufficiently large number of timesteps, the sum of these terms will approximate the amount of energy lost due to leakage in DES units. That is, taking an energy of $\Delta E$ from the grid using a DES unit $d \in \mathcal D_{DES}$ will yield a cost of $\Delta E$. Given a charging and discharging efficiency factor of $\eta_d$ for $d$, injecting the remaining energy after a total round-trip loss will result in a cost of $-\eta^2\Delta E$, totalling a round-trip cost of $(1-\eta^2)\Delta E$. This is the total energy loss over the round-trip.
    \item $\Delta E_{t:t+1}^{(3)}$ is the total amount of energy loss as a result of renewable generation curtailment of generators $\mathcal D_{RER}$ during $(t, t+1]$. Depending on the regulation, this can be thought of as a fee paid by the DNO to the owners of the generators that get curtailed, as financial compensation.
\end{itemize}

In the penalty term $\phi(s_{t+1})$, we consider two types of network-wide operating constraints. The first is the limit on the amount of power\footnote{In the literature, these limits are sometimes described in terms of current flows, instead of power flows.} that can flow through a transmission link $e_{ij} \in \mathcal E$, referred to as the rating of that link. These constraints are needed to prevent lines and transformers from overheating. The second type of constraint is a limit on the allowed voltage magnitude $|V_i|$ at each bus $i \in \mathcal N$. The latter are necessary conditions to maintain stability throughout the network and ensure proper operation of devices connected to the grid. 

In practice, violating any network constraint can lead to damaging parts of the DN infrastructure (e.g., lines or transformers) or power outages. Both can have important economic consequences for the DNO. For that reason, ensuring that the DN operates within its constraints is often prioritized compared to minimizing energy loss. Although our choice of reward function does not guarantee that an optimal policy will never violate these constraints, choosing a large $\lambda$ will ensure that these violations remain small. This would, in practice, have a negligible impact on the operation of the DN. In addition, the risk of violating real-life constraints in the DN could be further reduced by setting an over-restrictive set of constraints in the environment.

The technical details behind the computation of $r_t$ can be found in Appendix \ref{app:reward_function}.

\subsection{Model Predictive Control Scheme}
\label{sec:baseline}

In order to quantify how well an agent is performing on a specific Gym-ANM task, we can cast the task as a MPC problem in which a multi-stage ($N$-stage) OPF problem is solved at each timestep. The resulting policy provides us with a loose lower bound on the best performance achievable in the environment.

The general MPC algorithm that we provide takes as input forecasts of demand for each load $l \in \mathcal D_L$ and of maximum generation for each non-slack generator $g \in \mathcal D_G - \{g^{slack}\}$ over the optimization horizon $[t+1,t+N]$. We refer to the resulting policy as $\pi_{MPC-N}$. We then consider two variants: policies $\pi_{MPC-N}^{constant}$ and $\pi_{MPC-N}^{perfect}$. The former, $\pi_{MPC-N}^{constant}$, uses constant forecasts over the optimization horizon. Its simplicity means that it can be used in any Gym-ANM environment\footnote{See the project repository for more information.}. The other variant, $\pi_{MPC-N}^{perfect}$, assumes perfect predictions of future demand and generation are available for planning. Although it can only be used in simple environments such as ANM6-Easy (see Section \ref{sec:anm6-easy}), its performance is superior to that of $\pi_{MPC-N}^{constant}$. This means it provides the user with a tighter lower bound on the best achievable performance. Both variants are formally described in Appendix \ref{app:opf}.

Both MPC-based control schemes model the power grid using the DC power flow equations, a linearized version of the AC power flow equations. They thus solve a multi-stage DCOPF problem at each timestep. The DCOPF formulation relies on three assumptions: (a) transmission lines are lossless, (b) the difference between adjacent bus voltage angles is small, and (c) bus voltage magnitudes are close to unity. 

It is worth stressing that the MPC method that we propose here is an example of a traditional approach to tackling ANM problems. Because RL algorithms make less assumptions about the intrinsic structure of the problem, however, they have the potential to overcome the limitations of such optimization approaches and reach better solutions. This, of course, does not mean that RL should be blindly applied to most multi-step OPF-like problems, but, rather, that it might prove to be a good alternative when traditional approaches reach their limitations. This remains a hypothesis which, we hope, Gym-ANM will help confirm or deny.

\section{Environments}
\label{sec:environments}

\subsection{Gym-ANM Environments.} In conformity with the Gym framework, any Gym-ANM environment provides four main functions that allow the agent to interact with it: \texttt{reset()}, \texttt{step(action)}, \texttt{render()}, and \texttt{close()}. An example of code illustrating the interactions between an agent \texttt{agent} and an environment \texttt{env} is shown in Listing \ref{lst:code_snippet} (inspired from \cite{brockman2016openai}). The agent-learning procedure is omitted for clarity. Guidelines to design and implement new Gym-ANM environments can be found in Appendix \ref{app:new_env}.

{\footnotesize
\begin{lstlisting}[language=Python, caption={A code snippet (Python 3) illustrating environment-agent interactions.}, label=lst:code_snippet, captionpos=b, escapeinside={(*}{*)}]
env = gym.make('MyANMEnv')   # Initialize the environment.
obs = env.reset()            # Reset the env. and collect (*$o_0$*).

for t in range(1, T):         
  env.render()         # Update the rendering.
  a = agent.act(obs)   # The agent takes (*$o_t$*) as input and chooses (*$a_t$*).
  obs, r, done, info = env.step(a) 
        # The action is applied, and are outputted:
        # - obs: the new observation (*$o_{t+1}$*),
        # - r: the reward (*$r(s_t, a_t, s_{t+1})$*),
        # - done: True if (*$s_{t+1} \in \mathcal S^{terminal}$*),
        # - info: extra info about the transition.

env.close()  # Close the environment and stop rendering.
\end{lstlisting}} 

\subsection{ANM6-Easy.} 
\label{sec:anm6-easy}

Along with this paper we are also releasing ANM6-Easy, a Gym-ANM environment that models a series of ANM characteristic problems. ANM6-Easy is built around a DN consisting of six buses, with one high-voltage to low-voltage transformer, connected to a total of three passive loads, two renewable energy generators, one DES unit, and one fossil fuel generator used as slack generator. The topology of the network is shown in Figure \ref{fig:env_example} and its technical characteristics are summarized in Appendix \ref{app:smartgrid-env6_specs}. We use a time discretization of $\Delta t = 0.25$ (i.e., 15 minutes) by analogy with the typical duration of a market period, much like the work of \cite{gemine2017active}. The \texttt{observation()} component is the identity function. This leads to a fully observable environment with $o_t = s_t$. The discount factor is fixed to $\gamma = 0.995$, the reward penalty to $\lambda = 10^3$, and the reward clipping value to $r_{clip}=100$.

In order to limit the complexity of the task, we also chose to make the processes generated by the \texttt{next\_vars()} block deterministic. To do so, we use a fixed 24-hour time series that repeats every day, indefinitely. A single auxiliary variable $aux_t^{(0)} = (T_0+t) \mod{\frac{24}{\Delta t}}$ representing the time of day is used to index the time series, where $T_0 \in \{0,1, \ldots,\frac{24}{\Delta t}-1\}$ is the starting timestamp of the trajectory. During each timestep transition, the \texttt{next\_vars()} function thus behaves as described by Algorithm \ref{algo:next_vars}, where $\boldsymbol P_l[0,\ldots,\frac{24}{\Delta t}-1]$ and $\boldsymbol P_g[0,\ldots,\frac{24}{\Delta t}-1]$ are the fixed daily time series of load injections $P_{l,t}^{(dev)}$ and maximum generations $P_{g,t}^{(max)}$, respectively. The initialization procedure of the environment is also provided in Appendix \ref{app:smartgrid-env6_specs}.

\begin{algorithm}
\setstretch{1.2}
\caption{Implementation of \texttt{next\_vars()} in ANM6-Easy.}
\label{algo:next_vars}
\begin{algorithmic}[1]
\STATE $aux_{t+1}^{(0)} \gets (aux_{t}^{(0)} + 1) \mod{\frac{24}{\Delta t}}$
\FOR{$l \in \mathcal D_L$}
\STATE $P_{l,t+1}^{(dev)} \gets \boldsymbol P_l[aux_{t+1}^{(0)}]$
\ENDFOR 
\FOR{$g \in \mathcal D_G - \{g^{slack}\}$}
\STATE $P^{(max)}_{g,t+1} \gets \boldsymbol P_g[aux_{t+1}^{(0)}]$
\ENDFOR 
\end{algorithmic}
\end{algorithm}

\begin{figure}[h]
    \centering
    \fbox{\includegraphics[width=.9\linewidth]{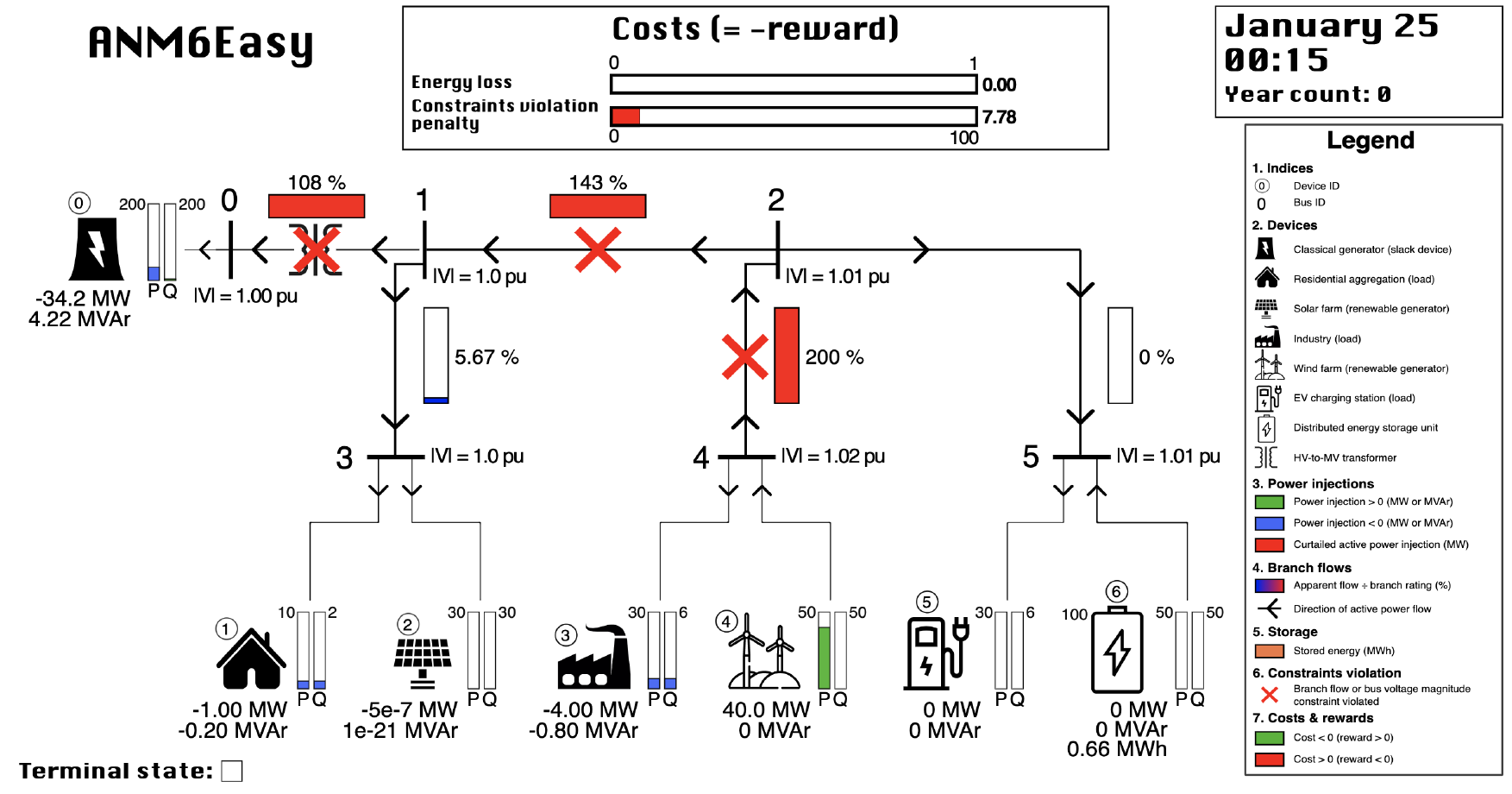}}
    \caption{Situation 1, lasting between 11:00 p.m. and 06:00 a.m. every day.}
    \label{fig:env6_situation1}
\end{figure}

\begin{figure}[h]
    \centering
    \fbox{\includegraphics[width=.9\linewidth]{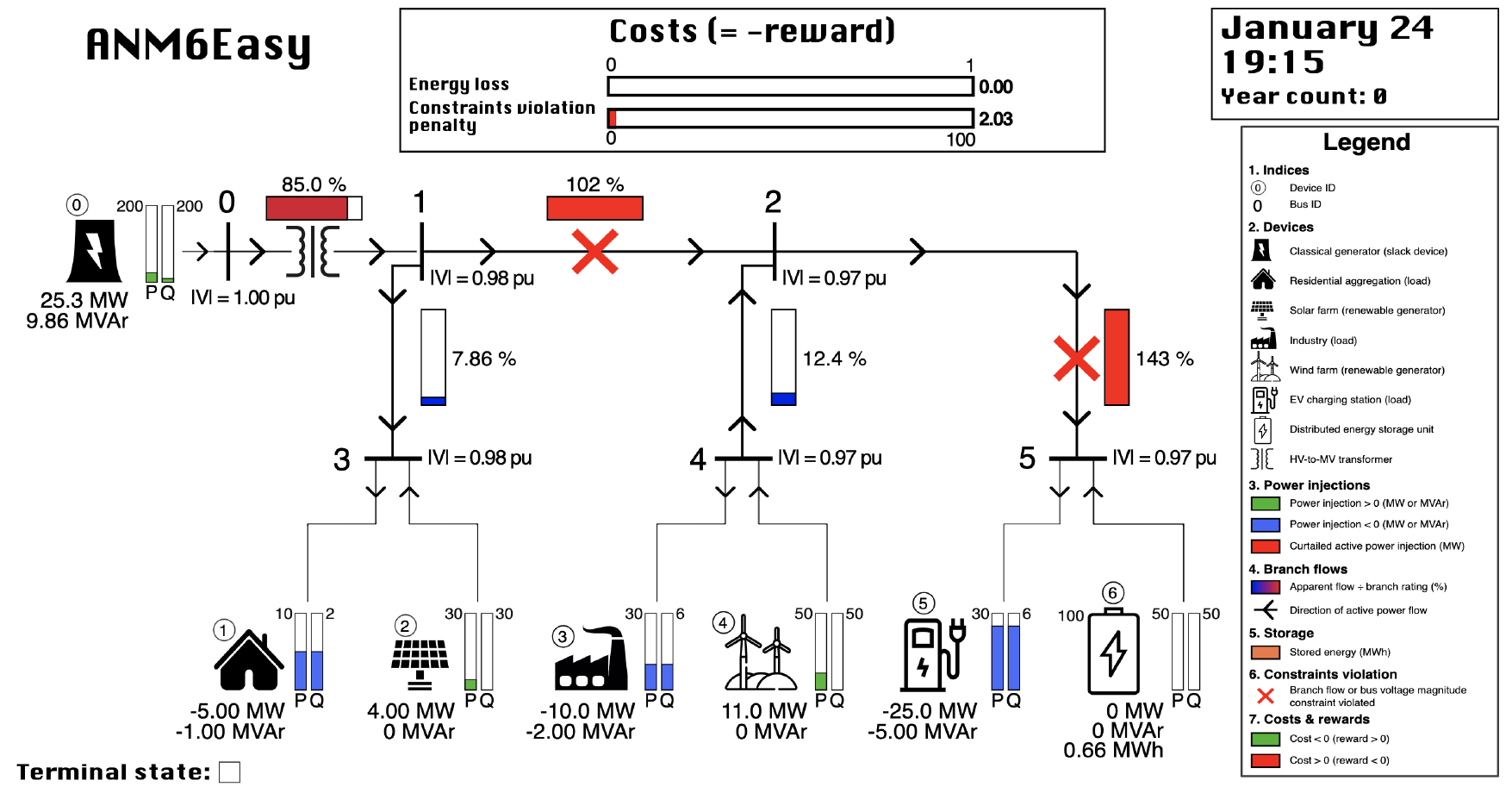}}
    \caption{Situation 2, lasting between 08:00 a.m. and 11:00 a.m. and between 06:00 p.m. and 09:00 p.m. every day.}
    \label{fig:env6_situation2}
\end{figure}

\begin{figure}[h]
    \centering
    \fbox{\includegraphics[width=.9\linewidth]{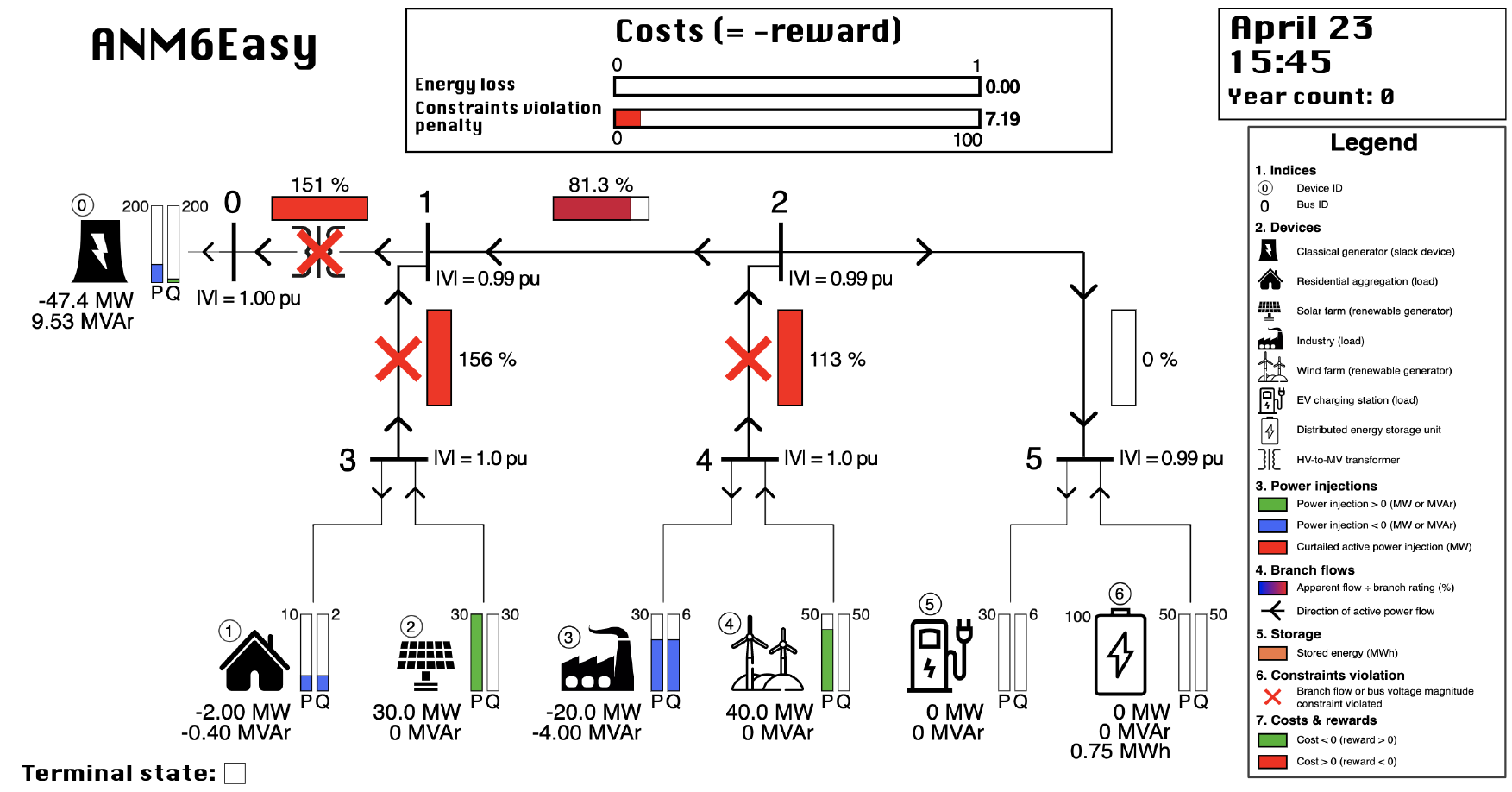}}
    \caption{Situation 3, lasting between 01:00 p.m. and 04:00 p.m. every day.}
    \label{fig:env6_situation3}
\end{figure}

The daily patterns were engineered so as to produce three problematic situations in the DN. Figures \ref{fig:env6_situation1}, \ref{fig:env6_situation2}, and \ref{fig:env6_situation3} show the power injections, power flows, and voltage levels that would result in each situation if the agent neither curtailed the renewable energies nor used the DES unit. Each situation lasts for seven, three, and three hours, respectively, during which the power injections remain constant. A two-hour-long period is used to transition between situations, during which each power injection is linearly incremented from its old to new value.

\paragraph{Situation 1} This situation (Figure \ref{fig:env6_situation1}) characterizes a windy night, when the consumption is low, the PV production null, and the wind production at its near maximum. Due to the very low demand from the industrial load, the wind production must be curtailed to avoid an overheating of the transmission lines connecting buses 0 and 4. This is also a period during which the agent might use this extra generation to charge the DES unit in order to prepare to meet the large morning demand from the EV charging garage (see Situation 2).

\paragraph{Situation 2} In this situation (Figure \ref{fig:env6_situation2}), bus 5 is experiencing a substantial demand due to a large number of EVs being plugged-in at around the same time. This could happen in a large public EV charging garage. In the morning, workers of close-by companies would plug in their car after arriving at work and, in the evening, residents of the area would plug in their cars after getting home. In order to emphasize the problems arising from this large localized demand, we assume that the other buses (3 and 4) inject or withdraw very little power into/from the network. During those periods of the day, the DES unit must provide enough power to ensure that the transmission path from bus 0 to bus 5 is not over-rated, which would lead to an overheating of the line. For this to be possible, the agent must strategically plan ahead to ensure a sufficient charge level at the DES unit.

\paragraph{Situation 3} Situation 3 (Figure \ref{fig:env6_situation3}) represents a scenario that might occur in the middle of a sunny windy weekday. No one is home to consume the solar energy produced by residential PVs at bus 1 and the wind energy production exceeds the industrial demand at bus 2. In this case, both renewable generators should be adequately curtailed while again storing some of the extra energy to anticipate the EV late afternoon charging period, as depicted in Situation 2.

\section{Experiments}
\label{sec:experiments}
In this section, we illustrate the use of the Gym-ANM framework. We compare the performance of PPO and SAC, two model-free deep RL algorithms, against that of the MPC-based policies $\pi_{MPC-N}^{constant}$ and $\pi_{MPC-N}^{perfect}$ introduced in Section \ref{sec:baseline} on the ANM6-Easy task. For both algorithms, we used the implementations from Stable Baselines 3 \cite{stable-baselines3}, a popular library of RL algorithms. Since our goal was not to compute an excellent approximation of an optimal policy, but rather to show that existing RL algorithms can already yield good performance with very little hyperparameter tuning, most hyperparameters were set to their default value (see Appendix \ref{app:hyperparameters}). The code used for all experiments in this section can be found at \url{https://github.com/robinhenry/gym-anm-exp}.

\subsection{Algorithms}
\label{sec:algorithms}

\paragraph{Proximal Policy Optimization} PPO is a stable and effective on-policy policy gradient algorithm. It alternates between collecting experience, in the form of finite-length trajectories starting from states $s_0 \sim p_0(\cdot)$ and following the current policy, and performing several epochs of optimization on the collected data to update the current policy (after which the collected experience is discarded). During each policy update step, the policy parameters $\theta$ are updated by maximizing (e.g., stochastic gradient ascent) a clipped objective function characterized by a hyperparameter $\epsilon$ that dictates how far away the new policy $\pi_{\theta}$ is allowed to diverge from the old $\pi_{\theta_{old}}$. The objective also requires the use of an advantage-function estimator, which is achieved using a learned-state value function $V_\phi(s)$. In the Stable Baselines 3 implementation that we used, both the policy $\pi_\theta$ and the state-value function $V_\phi(s)$ were represented using separate fully connected MLPs with weights $\theta$ and $\phi$, respectively, each with two layers of 64 units and tanh nonlinearities.

\paragraph{Soft Actor-Critic} SAC is an off-policy actor-critic algorithm based on the maximum entropy RL framework. The policy is trained to maximize a trade-off between expected return and entropy, a measure of randomness in the policy. It alternates between collecting and storing experience of the form $(s_t, a_t, r_t, s_{t+1})$ into a replay buffer, regularly ending the current trajectory to start from a new initial state $s_0 \sim p_0(\cdot)$, and updating the policy $\pi_\theta$ (actor) and a soft Q-function $Q_\phi(s_t, a_t)$ (critic) from batches sampled from the replay buffer (e.g., stochastic gradient descent), in an offline manner. In the same manner as the work of Haarnoja et al. \cite{haarnoja2018soft}, the implementation that we used makes use of two Q-functions to mitigate positive bias in the policy improvement step. Both the policy $\pi_{\theta}$ and the Q-functions $Q_{\phi_1}$, $Q_{\phi_2}$ were represented using separate fully connected MLPs with weights $\theta$, $\phi_1$, and $\phi_2$, respectively, each with two layers of 64 units and ReLU nonlinearities. Separate target Q-networks that slowly track $Q_{\phi_1}$, $Q_{\phi_2}$ were also used to improve stability, using an exponentially moving average with smoothing constant $\tau$.

\subsection{Performance metric}
We evaluate the performance of the different algorithms on the ANM6-Easy task as follows. Every $N_{eval}$ steps the agent takes in the environment (i.e., selects an action), we freeze the training procedure and evaluate the current policy on another instance of the environment. To do so, we collect $N_{r}$ rollouts of $T$ timesteps each, using the current policy $\pi_\theta$, and report:
\begin{align}
    J_{\pi_\theta} = \mathbb E_{s_0 \sim p_0(\cdot)}[J_{\pi_\theta}(s_0)] \approx \frac{1}{N_{r}} \sum_{i=1}^{N_{r}} \sum_{t=0}^{T-1} \gamma^t r_t^{(i)} ,
\end{align}
where $s_0^{(i)} \sim p_0(\cdot)$ and $r_t^{(i)}$ are the initial state and rewards obtained in the $i$\textsuperscript{th} rollout, respectively. Because the reward signal is bounded by a finite constant $r_{clip} \in \mathbb R$ (i.e., $|r_t| \in [-r_{clip}, r_{clip}]$, $\forall t$), approximating $J_{\pi_\theta}(s_0) = \lim_{T\to \infty} \sum_{t=0}^{T-1} \gamma^t r_t$ by $\sum_{t=0}^{T-1}\gamma^t r_t$ may result in a deviation of up to $r_{clip}\frac{\gamma^T}{1-\gamma}$ from the true infinite discounted return, since:
\begin{align}
J_{\pi_\theta}(s_0) \le r_{clip}\frac{1}{1-\gamma} \quad \text{and} \quad \sum_{t=0}^{T-1}\gamma^t r_t \le r_{clip} (\frac{1}{1-\gamma} - \frac{\gamma^T}{1-\gamma}) \;.
\end{align}
In our experiments, we used $N_r = 5$ and set $T=3000$, such that $r_{clip}\frac{\gamma^T}{1-\gamma} < 10^{-2}$ results in negligible error terms.

\subsection{Results}
\begin{figure}
    \centering
    \includegraphics[width=.85\textwidth]{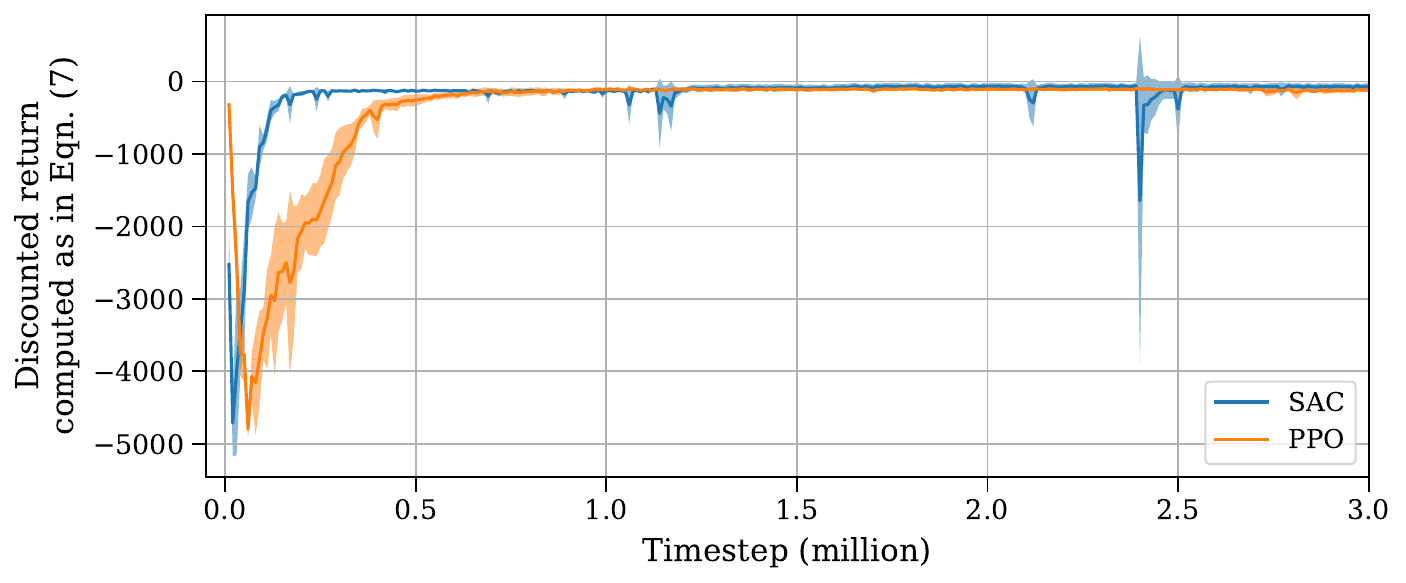}
    \caption{Evolution of the empirical discounted return $J_{\pi_\theta}$ ($T=3000$) during training.}
    \label{fig:discounted_returns}
\end{figure}

We trained both the PPO and SAC algorithms on the ANM6-Easy environment for three million steps, starting from a new initial state $s_0 \sim p_0(\cdot)$ every 5000 steps (or earlier if a terminal state is reached), and evaluated their performance every $N_{eval} = 10^4$ steps. Both algorithms used normalized observation and action vectors. We repeated the same procedure with 5 random seeds and plotted the mean and standard deviation of the evolution of their performance during training in Figure \ref{fig:discounted_returns}. 

Table \ref{tab:mpc_discounted_returns} reports the average performance of policies $\pi_{MPC-N}^{constant}$ and $\pi_{MPC-N}^{perfect}$ for different planning steps $N$ and safety margin hyperparameters $\beta$ (see Appendix \ref{app:opf}). As expected, the performance of $\pi_{MPC-N}^{perfect}$ increases with $N$, since the algorithm has access to perfect demand and generation forecasts. In the case of $\pi_{MPC-N}^{constant}$, the best average return is capped at 129.1 and increasing $N$ does not improve performance.

\begin{table}[!htb]
    \begin{minipage}{.45\linewidth}
      \centering
      \footnotesize
        \begin{tabular}{c|ccccc}
            \toprule
            $\beta$ \textbackslash $N$ & 8 & 16 & 32 \\
            \midrule
            0.92 & {\bf -129.1} & {\bf -129.1} & {\bf -129.1} \\
            0.94 & -129.3 & -129.3 & -129.2 \\
            0.96 & -129.6 & -129.5 & -129.5 \\
            0.98 & -130.5 & -130.5 & -130.5 \\
            1    & -134.8 & -134.7 & -134.7 \\
            \bottomrule
        \end{tabular}
    \end{minipage}%
    \begin{minipage}{.55\linewidth}
      \centering
      \footnotesize
        \begin{tabular}{c|ccccc}
            \toprule
            $\beta$ \textbackslash $N$ & 8 & 16 & 32 & 64 \\
            \midrule
            0.92 & -100.6 & -60.3 & -16.0 & -16.0 \\
            0.94 & -99.7  & -58.2 & {\bf -14.7} & {\bf -14.7} \\
            0.96 & -102.1 & -57.5 & -14.8 & -14.8 \\
            0.98 & -102.4 & -59.6 & -19.0 & -19.0 \\
            1    & -108.0 & -68.5 & -29.1 & -29.1 \\
            \bottomrule
        \end{tabular}
    \end{minipage} 
    \vspace{0.2cm}
    \caption{Average discounted returns $J_{\pi_\theta}$ for $\pi_{MPC-N}^{constant}$ (left) and $\pi_{MPC-N}^{perfect}$ (right), for different planning horizons $N$ and safety margin hyperparameters $\beta$.}
    \label{tab:mpc_discounted_returns}
\end{table}

Table \ref{tab:final_performance} compares the best performance of the trained agents against that of the MPC policies. Note that both RL agents reach better performances than $\pi_{MPC-N}^{constant}$. That is, both PPO and SAC outperform a MPC-based policy in which future demand and generation are assumed constant. 

Finally, Table 2 also summarizes computational CPU times required for each control policy to select an action on a MacBook 2.3 GHz Intel Core i5 with 8GB of RAM. Clearly, RL policies have the advantage of requiring significantly less time for action selection, since the mapping from state (or observation) to action is stored in the form of function approximators, which can be efficiently evaluated. Nevertheless, the learning of these function approximators may require significant computational times, which vary greatly between different RL algorithms.

\begin{table}[h]
    \centering
    \begin{tabular}{lcccc}
    \toprule
    & PPO & SAC & $\pi_{MPC-16}^{constant}$ &  $\pi_{MPC-32}^{perfect}$ \\
    \midrule
    $J_{\pi_{\theta}}$ & -93.6 $\pm$ 15.3 & -56.1 $\pm$ 26.8 & -129.1 $\pm$ 0.4 & -14.7 $\pm$ 0.2 \\
    Time (ms) & 0.47 $\pm$ 0.19 & 0.52 $\pm$ 0.28 & 31.60 $\pm$ 30.38 & 61.75 $\pm$ 31.21 \\
    \bottomrule
    \end{tabular}
    \vspace{0.2cm}
    \caption{Top row: mean and standard deviation of the best discounted returns over 5 random seeds. Bottom row: mean and standard deviation of the CPU time required to select an action on a MacBook 2.3 GHz Intel Core i5 with 8GB of RAM.}
    \label{tab:final_performance}
\end{table}

\section{Conclusion}
\label{sec:conclusion}
In this paper, we proposed Gym-ANM, a framework for designing and implementing RL environments that model ANM problems in electricity distribution networks. We also introduced ANM6-Easy, a particular instance of such environments that highlights common challenges in ANM. Finally, we showed that state-of-the-art RL algorithms can already reach performances similar to that of MPC-based policies that solve multi-stage DCOPF problems, with little hyperparameter tuning. 

We hope that our work will inspire others in the RL community to tackle decision-making problems in electricity networks, potentially through the use of our framework. We believe that Gym-ANM has the potential to model tasks of a wide range of complexity, creating a novel extensive playground for advanced RL research.

\section{Acknowledgements}
We would like to thank Raphael Fonteneau, Quentin Gemine, and Sébastien Mathieu at the University of Liège for their valuable early feedback and advice, as well as Gaspard Lambrechts and Bardhyl Miftari for the feedback they provided as the first users of Gym-ANM.

\bibliography{references}

\newpage
\begin{appendices}

\section{Electricity Distribution Network Simulator}
\label{app:simulator}
This Appendix describes in more detail the dynamics of the alternative current (AC) power grid on top of which Gym-ANM environments are built. Section \ref{app:preliminaries} introduces some technical power system notions used in later analyses. Sections \ref{app:branch_model}, \ref{app:passive_loads}, \ref{app:distributed_generators}, and \ref{app:des} describe the mathematical model and assumptions used to simulate the behavior of transmission links, passive loads, distributed generators, and DES units, respectively. Section \ref{app:network_constraints} then introduces the set of network constraints that we would like the learned ANM control scheme to satisfy, and Section \ref{app:network_equations} derives the set of equations that govern the network electricity flows. Finally, Sections \ref{app:transition_function} and \ref{app:reward_function} derive the sequence of computational steps that make up the environment transition and reward functions, respectively.

\subsection{Preliminaries}
\label{app:preliminaries}
Today, the majority of AC transmission and distribution networks dispatch electricity using the so-called three-phase system. In this system, electricity flows in three parallel circuits, each associated with its own phase. In a balanced three-phase network, the electrical quantities of each phase have the same magnitude and differ by a $120\degree$ phase shift, i.e. phase 3 is a time-delayed version of phase 2, which is itself a time-delayed version of phase 1. Conveniently, any balanced three-phase system can thus be analyzed using an equivalent single-phase representation, where only one of the phases is taken into account. The complex phasors corresponding to the other two phases can be obtained by applying a 120$\degree$ or 240$\degree$ phase shift to the first-phase phasors. All systems implemented by Gym-ANM are assumed to be such three-phase balanced networks, and we adopt its equivalent single-phase representation in the following derivations.

In order to efficiently generate and distribute electricity, power grids are also divided into so-called voltage zones. Each zone is characterized by a particular nominal voltage level that represents the average voltage level of the nodes in that zone. For instance, a 220kV (ultra-high voltage) transmission network may be connected to an intermediary 150kV (high voltage) network, which is then connected to a 30kV (medium voltage) distribution network. Transitions between the different voltage levels are carried out by power transformers that bring up (step-up transformers) or down (step-down transformers) voltages while minimizing power losses. For mathematical convenience, power systems that include several voltage zones are often analyzed using the per-unit (p.u.) notation, in which all electrical quantities are normalized with respect to a set of base quantities chosen for the whole system. In practice, the per-unit analysis method becomes very handy as it removes the need to include nominal voltage levels in derivations. This allows us to analyze the network as a single circuit and cancels out the effect of transformers whose tap ratio is identical to the ratio of the base voltages of the zones it connects. In other words, only so-called off-nominal transformers need to be considered. In the remainder of this Appendix, all quantities are expressed in p.u.

\subsection{Branches}
\label{app:branch_model}
\begin{figure}[h]
    \centering
    \includegraphics[width=0.85\linewidth]{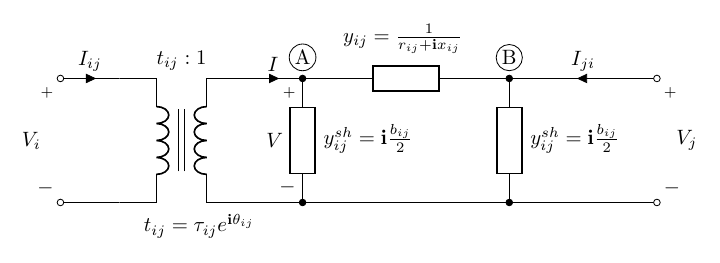} \caption{Common branch model, consisting of a $\pi$ transmission line model in series with an off-nominal phase-shifting transformer, connecting bus $i \in \mathcal N$ and bus $j \in \mathcal N$.}
    \label{fig:pi_model}
\end{figure}

As introduced in Section \ref{sec:background_dn}, we model a distribution network as a set of nodes $\mathcal N$ connected by a set of directed edges $\mathcal E$. Each edge $e_{ij} \in \mathcal E$ may represent a sequence of (a) transmission lines, (b) power transformers, and/or (c) phase shifters linking buses $i$ and $j$. Any combination of (a)-(c) components can be equivalently mapped to the common branch representation adapted from \cite{zimmerman2010matpower} and shown in Figure \ref{fig:pi_model}. Formally, branch $e_{ij} \in \mathcal E$ is characterized by five parameters: a series resistance $r_{ij}$, a series reactance $x_{ij}$, a total charging susceptance $b_{ij}$, a tap ratio magnitude $\tau_{ij}$, and a phase shift $\theta_{ij}$. The branch series admittance is given by $y_{ij} = (r_{ij} + \mathbf i x_{ij})^{-1}$, each shunt admittance by $y_{ij}^{sh} = \mathbf i \frac{b_{ij}}{2}$, and the complex tap ratio of the off-nominal transformer by $t_{ij} = \tau_{ij}e^{\mathbf i \theta_{ij}}$. Note that one can use a value of $t_{ij} = 1$ to represent the absence of a transformer, or, equivalently, the presence of an on-nominal transformer.

\subsection{Electrical Devices}
\label{app:electrical_devices}
The different electrical devices $\mathcal D$ connected to the grid are classified as passive loads $\mathcal D_L$, generators $\mathcal D_G$, or DES units $\mathcal D_{DES}$. Within generators, we further differentiate between renewable generators $\mathcal D_{RER} \subset \mathcal D_G$ and the slack generator $g^{slack} \in \mathcal D_G - \mathcal D_{RER}$. Much like what was done by Gemine et al.\cite{gemine2017active}, the range of operation of each device $d \in \mathcal D$ is modelled by a set $\mathcal{R}_{d,t} \subset \mathbb R^2$ of valid $(P_{d,t}^{(dev)}, Q_{d,t}^{(dev)})$ power injection points for timestep $t$. These constraints are enforced by the environment at all times (see Appendix \ref{app:transition_function}).

\subsubsection{Passive Loads}
\label{app:passive_loads}
We define passive loads as the devices that only withdraw power from the network. We also assume that each passive load $l \in \mathcal D_L$ has a constant power factor $\cos{\phi_l}$ and that its negative injection $P_{l,t}^{(dev)}$ is lower bounded\footnote{When designing a new environment, the user can set $\underbar{P}_l = - \infty$ to model an unbounded load. Note that a finite lower bound value is required to have a bounded state space $\mathcal S$.} by $\underline P_l$. Formally, the range of operation $\mathcal R_{l,t} = \mathcal R_l$ of $l$ is defined by:
\begin{align}
    \mathcal R_l = \{(P, Q) \in \mathbb R^2 \; | \; \underline P_l \le P \le 0,
    \frac{Q}{P} = \tan{\phi_l} \}\;, \quad \forall l \in \mathcal D_L \;, \label{eq:restriction_load_injection_points}
\end{align}
for all $t \in \mathcal T$.

\subsubsection{Generators}
\label{app:distributed_generators}
\begin{figure}[h]
    \centering
    \includegraphics[width=0.5\linewidth]{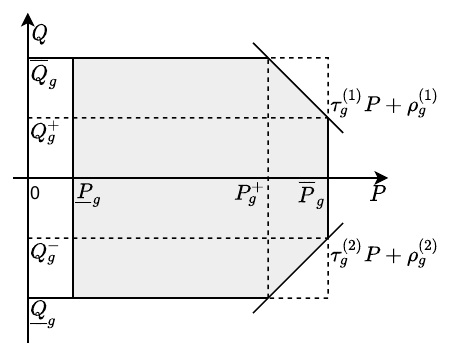}
    \caption{Fixed power injection constraints of distributed generators $g \in \mathcal D_G - \{g^{slack}\}$.}
    \label{fig:dg_polyhedron}
\end{figure}

Generators, with the exception of $g^{slack}$, refer to devices that only inject power into the network. The physical limitations of any generator $g \in \mathcal D_G$ are modelled by a range of allowed active power injections $[\underline P_g, \overline P_g]$ and of reactive power injections $[\underline Q_g, \overline Q_g]$. Additional linear constraints $Q_{g,t}^{(dev)} \le \tau_g^{(1)} P_{g,t}^{(dev)} + \rho_g^{(1)}$ and $Q_{g,t}^{(dev)} \ge \tau_g^{(2)} P_{g,t}^{(dev)} + \rho_g^{(2)}$ can also be added to limit the flexibility of reactive power injection when $P$ is close to its maximum value.\footnote{These additional linear flexibility constraints can be used to approximate current limits of power converters and/or of electric generators \cite{engelhardt2010reactive}. They can also be ignored by setting $Q_g^+ = \overline Q_g$ and $Q_g^- = \underline Q_g$.} These constraints result in the range of operation shown in Figure \ref{fig:dg_polyhedron}. Finally, a dynamic upper bound $P^{(max)}_{g,t} \in [\underline P_g, \overline P_g]$ is also generated by the \lstinline{next_vars()} block to model time-dependent constraints on $P_{g,t}^{(dev)}$.

The resulting dynamic region of operation $\mathcal R_{g,t}$ is formally expressed as: 
\begin{align}
    \mathcal R_{g,t} = \{(P, Q) \in \mathbb R^2 \; | \; 
        & \underline P_g \le P \le P^{(max)}_{g,t} \;, \nonumber \\ 
        & \underline Q_g \le Q \le \overline{Q}_g \;, \nonumber \\
        & Q \le \tau_g^{(1)} P + \rho_g^{(1)} \;, \nonumber \\
        & Q \ge \tau_g^{(2)} P + \rho_g^{(2)}\} \;, \quad \forall g \in \mathcal D_G - \{g^{slack}\} \;, \label{eq:dg_polyhedron}
\end{align}
where $\tau_g^{(1)}, \rho_g^{(1)}, \tau_g^{(2)}, \rho_g^{(2)}$ are computed based on the parameters $\{\overline P_g, P^+_g, \underline Q_g,$ $\overline Q_g, Q_g^+, Q_g^-\}$ provided in the network input dictionary (see Appendix \ref{app:case_file}) as:
\begin{align}
    & \tau_g^{(1)} = \frac{Q^+_g - \overline Q_g}{\overline P_g - P^+_g}, \quad \rho_g^{(1)} = \overline Q_g - \tau_g^{(1)} P^+_g, \nonumber \\
    & \tau_g^{(2)} = \frac{Q^-_g - \underline Q_g}{\overline P_g - P^+_g}, \quad \rho_g^{(2)} = \underline Q_g - \tau^{(2)}_g P^+_g.
\end{align}

In order to ensure that a solution to the network equations derived in Section \ref{app:network_equations} is found at each timestep, we do not restrict the range of operation of the slack generator $g^{slack}$. Instead, we assume that it can provide unlimited active and reactive power to the network.

\subsubsection{Distributed Energy Storage (DES)}
\label{app:des}
\begin{figure}[h]
    \centering
    \includegraphics[width=0.7\linewidth]{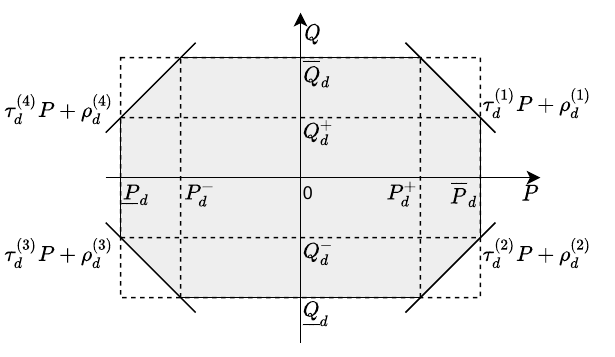}
    \caption{Fixed power injection constraints for DES units $d \in \mathcal D_{DES}$.}
    \label{fig:des_valid_inj_points}
\end{figure}

DES units can both inject power into (discharge) and withdraw power from (charge) the network. Their time-independent physical constraints are modelled much like that of generators, as shown in Figure \ref{fig:des_valid_inj_points}, where:
\begin{align}
    &\tau_d^{(1)} = \frac{Q^+_d - \overline Q_d}{\overline P_d - P^+_d}, \quad
    \rho_d^{(1)} = \overline Q_d - \tau_d^{(1)} P^+_d, \\
    & \tau_d^{(2)} = \frac{Q^-_d - \underline Q_d}{\overline P_d - P^+_d}, \quad
    \rho_d^{(2)} = \underline Q_d - \tau^{(2)}_d P^+_d, \\
    & \tau_d^{(3)} = \frac{\underline Q_d - Q^-_d}{P_d^- - \underline P_d}, \quad
    \rho_d^{(3)} = \underline Q_d - \tau_d^{(3)} P^-_d, \\
    & \tau_d^{(4)} = \frac{\overline Q_d - Q^+_d}{P_d^- - \underline P_d}, \quad
    \rho_d^{(4)} = \overline Q_d - \tau^{(4)}_d P^-_d.
\end{align}

Unlike generators, however, the active power injection of a DES unit $d \in \mathcal D_{DES}$ is further constrained by its current state of charge $SoC_{d,t} \in [\underline{SoC}_d, \overline{SoC}_d]$. For instance, a fully charged unit would not be able to withdraw even the slightest amount of active power. Consequently, we chose to impose additional limits on their next active power injection $P_{d,t+1}^{(dev)}$. This is to ensure that the injection can stay constant within $(t, t+1]$ without violating any storage level constraints, i.e. that $\underline{SoC}_{d} \le SoC_{d,t+1} \le \overline{SoC}_{d}$. Given that $SoC_{d,t+1}$ is obtained from:
\begin{align}
    SoC_{d,t+1} = 
    \begin{cases}
    SoC_{d,t} - \Delta t \eta P_{d,t+1}^{(dev)} \;, & \text{if } P_{d,t+1}^{(dev)} \le 0 \;, \\
    SoC_{d,t} - \frac{\Delta t}{\eta} P_{d,t+1}^{(dev)} \;, & \text{else,}
    \end{cases} \label{eq:next_soc}
\end{align}
where $\eta \in [0, 1]$ is the charging and discharging efficiency factor (assumed equal), the condition $\underline{SoC}_{d} \le SoC_{d,t+1} \le \overline{SoC}_{d}$ can be re-expressed as a constraint on $P_{d,t+1}^{(dev)}$ as:
\begin{align}
    \frac{1}{\Delta t \eta}(SoC_{d,t} - \overline{SoC}_d) \le P_{d,t+1}^{(dev)} \le \frac{\eta}{\Delta t}(SoC_{d,t} - \underline{SoC}_d) \;.
\end{align}

In summary, the range of operation of each DES unit $d \in \mathcal D_{DES}$ is modelled by the time-varying constrained set $\mathcal R_{d,t}$:
\begin{align}
    \mathcal R_{d,t} = \{(P, Q) \in \mathbb R^2 \; | \; 
        & \underline{P}_d \le P \le \overline{P}_d \;, \nonumber \\ 
        & \underline{Q}_d \le Q \le \overline{Q}_d \;, \nonumber \\
        & Q \le \tau_d^{(1)} P + \rho_d^{(1)} \;, \nonumber \\
        & Q \ge \tau_d^{(2)} P + \rho_d^{(2)} \;, \nonumber \\
        & Q \ge \tau_d^{(3)} P + \rho_d^{(3)} \;, \nonumber \\
        & Q \le \tau_d^{(4)} P + \rho_d^{(4)} \;, \nonumber \\
        & P \ge \frac{1}{\Delta t \eta}(SoC_{d,t-1} - \overline{SoC}_d) \;, \\
        & P \le \frac{\eta}{\Delta t}(SoC_{d,t-1} - \underline{SoC}_d) \} \;, \quad \forall d \in \mathcal D_{DES}\; . \label{eq:storage_polyhedron}
\end{align}

\subsection{Network Constraints}
\label{app:network_constraints}
Constraints on the operating range of each electrical device in $\mathcal D$ (derived in Appendix \ref{app:electrical_devices}) get enforced by the environment during each timestep transition (see Appendix \ref{app:transition_function}). Unlike these constrains, however, network constraints will be left unchecked but will generate a large negative reward when not met, as further detailed in Appendix \ref{app:reward_function}. That is, the simulator will allow the network to operate past the following network constraints, but will penalize through negative rewards any policy that does so. 

As introduced in Section \ref{sec:reward_function}, we consider two types of such network constraints that network operators should ensure are satisfied at all times: voltage and line current constraints. The first one is a constraint on bus voltage magnitudes, which must be kept within a close range of their nominal value to ensure stability of the grid:
\begin{align}
    \underbar{V}_i \le |V_{i,t}| \le \overline{V}_{i} \;, \quad \forall i \in \mathcal N, \forall t \in \mathcal T  \;, \label{eq:voltage_constraint}
\end{align}
where $\underbar{V}_i$ and $\overline{V}_i$ are often chosen close to 1 p.u. and voltages are expressed as root mean squared (RMS) values.

The second one is an upper limit on line currents, which are determined by materials and environmental conditions. Let $\overline I_{ij}$ be the maximum physical current magnitude allowed through branch $e_{ij} \in \mathcal E$. In practice, such limits are often expressed as apparent power flow limits $\overline S_{ij}$ at a 1 p.u. nodal voltage. The reason behind this choice is the fact that the apparent power flow $|S_{ij}| = |V_i I_{ij}^*|$ is close to $|I_{ij}|$ when voltage magnitudes are kept close to unity by constraint (\ref{eq:voltage_constraint}). For consistency with existing optimization tools that model line current limits as apparent power flow constraints, we chose to adopt the same approach in Gym-ANM. In addition, for a given branch $e_{ij} \in \mathcal E$, the branch current at the sending end $|I_{ij}|$ may be different to the current injection at the receiving end $|I_{ji}|$. This is due to the asymmetry of the common branch model of Figure \ref{fig:pi_model}. The constraints must thus be respected at each end of the branch:
\begin{align}
    |S_{ij,t}| \le \overline{S}_{ij} \quad \text{and} \quad |S_{ji,t}| \le \overline{S}_{ij} \;, \quad \forall e_{ij} \in \mathcal E, \forall t \in \mathcal T \;. \label{eq:current_constraints}
\end{align}

\subsection{Network Equations}
\label{app:network_equations}
The flow of electricity within a power network is dictated by a set of network equations, or power flow equations, which we will now derive. The following derivations assume that all AC quantities are expressed in RMS terms.

The ideal transformer with complex tap ratio $t_{ij}:1$ used in the common branch model introduced in Section \ref{app:branch_model} can be further described by the relations:
\begin{align}
    V = \frac{V_i}{t_{ij}} \quad \text{and} \quad I = t_{ij}^*I_{ij} \;. \label{eq:transformer}
\end{align}
Applying Kirchhoff's current law at nodes A and B of Figure \ref{fig:pi_model} yields:
\begin{align}
    \begin{cases}
        I = Vy^{sh}_{ij} + (V-V_j)y_{ij} \\
        I_{ji} = V_j y_{ij}^{sh} + (V_j - V) y_{ij}
    \end{cases} \;,
\end{align}
which, after substituting (\ref{eq:transformer}), becomes:
\begin{align}
    \begin{cases}
         I_{ij} = \frac{1}{|t_{ij}|^2}(y_{ij} + y_{ij}^{sh})V_i - \frac{1}{t_{ij}^*}y_{ij}V_j \\
        I_{ji} = -\frac{1}{t_{ij}} y_{ij} V_i + (y_{ij} + y_{ij}^{sh})V_j
    \end{cases} \label{eq:current_injections} \;.
\end{align}
Expressions (\ref{eq:current_injections}) can be equivalently presented in matrix form:
\begin{align}
\begin{bmatrix}
        I_{ij} \\ I_{ji}
    \end{bmatrix}
    =
    \begin{bmatrix}
        \frac{1}{|t_{ij}|^2} (y_{ij} + y_{ij}^{sh}) &  - \frac{1}{t^*_{ij}} y_{ij} \\
        -\frac{1}{t_{ij}} y_{ij} & (y_{ij} + y_{ij}^{sh})
    \end{bmatrix}
    \begin{bmatrix}
        V_i \\ V_j
    \end{bmatrix} \;, \label{eq:branch_currents}
\end{align}
which is one possible formulation of the power flow equations.

However, the most commonly used formulation in practice is obtained after applying Kirchhoff's current law at each bus $i \in \mathcal N$, which results in the classical matrix formulation:
\begin{align}
    \mathbf I = \mathbf Y \mathbf V \;, \label{eq:pfe_matrix}
\end{align}
where $\mathbf I = [I_0, I_1, \ldots, I_{|\mathcal N|-1}]^T$ is the vector of bus current injections, $\mathbf V = [V_0, V_1, \ldots, V_{|\mathcal N|-1}]^T$ the vector of corresponding bus voltages, and $\mathbf Y \in \mathbb C^{|\mathcal N| \times |\mathcal N|}$ the nodal admittance matrix with elements:
\begin{align}
    \mathbf Y_{ij} = 
    \begin{cases}
        - \frac{1}{t_{ij}^*} y_{ij} \;, & \text{if } i \ne j \text{ and } e_{ij} \in \mathcal E \;, \\
        - \frac{1}{t_{ji}} y_{ji} \;, & \text{if } i \ne j \text{ and } e_{ji} \in \mathcal E \;, \\
        \sum_{e_{ik} \in \mathcal E} \frac{1}{|t_{ik}|^2} (y_{ik} + y_{ik}^{sh}) + \sum_{e_{ki} \in \mathcal E} (y_{ki} + y_{ki}^{sh}) \;, & \text{if } i = j \;, \\
        0 \;, & \text{otherwise} \;.
    \end{cases}
\end{align}
Finally, (\ref{eq:pfe_matrix}) can also be formulated in terms of nodal power injections and voltage levels, removing the need to compute current injections:
\begin{align}
    P_i^{(bus)} + \mathbf i Q_i^{(bus)} = V_i I_i^* = V_i (\mathbf Y_i \mathbf V)^* = V_i \mathbf Y_i^* \mathbf V^* \;, \quad \forall i \in \mathcal N \;, \label{eq:power_flow_equations}
\end{align}
where $\mathbf Y_i$ denotes the $i^{th}$ row of the admittance matrix $\mathbf Y$. 

The power flow equations (\ref{eq:power_flow_equations}) represent a set of $|\mathcal N|$ complex-valued equations that the environment solves during the \lstinline{next_state()} call of each timestep transition. To do so, every bus is modelled as a PQ bus: the $P_i^{(bus)}$ and $Q_i^{(bus)}$ variables are set by the environment (based on the agent's action) and the $V_i$ variables are left as free variables for the solver. The only exception is the slack bus, where the opposite is true: $V_i$ is fixed to $1\angle 0\degree$ and $P_i^{(bus)}$, $Q_i^{(bus)}$ are the variables. This setup results in a system of $2|\mathcal N|$ quadratic real-valued equations with $2|\mathcal N|$ free real variables.

\subsection{Transition Function}
\label{app:transition_function}
Based on the current state $s_t \in \mathcal S$, each timestep transition starts by sampling the internal variables through the \lstinline{next_vars()} block of Figure \ref{fig:env_structure}. Note that this block can be uniquely designed for different environments. The remainder of the transition function happens with the \lstinline{next_state()} component in a deterministic manner, which we now describe as a series of steps analogous to the underlying implementation.

\paragraph{1. Load injection point} First, the reactive power injection $Q_{l,t+1}^{(dev)}$ of each load $l \in \mathcal D_L$ is inferred from its new demand $P_{l,t+1}^{(dev)}$ outputted by \lstinline{next_vars()}, according to (\ref{eq:restriction_load_injection_points}):
\begin{align}
    Q_{l,t+1}^{(dev)} = P_{l,t+1}^{(dev)} \tan{\phi_l} \;,
\end{align}
where $P_{l,t+1}^{(dev)}$ is first clipped to $[\underline P_{l}, 0]$.

\paragraph{2. Distributed generator injection point} The power injection point of each distributed generator $g \in \mathcal D_G - \{g^{slack}\}$ is computed based on its allowed range of operation $\mathcal R_{g,t+1}$ given by (\ref{eq:dg_polyhedron}). The active and reactive injections $a_{P_{g,t}}$, $a_{Q_{g,t}}$ are then set by the agent in $a_t$: 
\begin{align}
    (P_{g,t+1}^{(dev)}, Q_{g,t+1}^{(dev)}) = \argmin_{(P, Q) \in \mathcal R_{g,t+1}} || \big(a_{P_{g,t}}, a_{Q_{g,t}} \big) - (P, Q)|| \;.
\end{align}
In the case where the $(a_{P_{g,t}}, a_{Q_{g,t}})$ injection point set by the agent falls outside of $\mathcal R_{g,t+1}$, the environment selects the closest point in $\mathcal R_{g,t+1}$, according to the Euclidean distance.

\paragraph{3. DES injection point} Similarly, the power injection point of each DES unit $d \in \mathcal D_{DES}$ is computed based on the $(a_{P_{d,t}}, a_{Q_{d,t}})$ point chosen by the agent in $a_t$ and the operating range $\mathcal R_{d,t+1}$ of $d$ given by (\ref{eq:storage_polyhedron}). We again use the Euclidean distance as the distance metric, resulting in:
\begin{align}
    (P_{d,t+1}^{(dev)}, Q_{d,t+1}^{(dev)}) = \argmin_{(P, Q) \in \mathcal R_{d,t+1}} \quad || (a_{P_{d,t}}, a_{Q_{d,t}}) - (P, Q) || \;.
\end{align}

\paragraph{4. Power flows \& bus voltages} Now that the power injection point of each device, with the exception of the slack generator, is known, the total nodal active and reactive power injection for each non-slack bus $i$ is computed using:
\begin{align}
    P_i = \sum_{d \in \mathcal D_i} P_d \quad \text{and} \quad Q_i = \sum_{d \in \mathcal D_i} Q_d \;.
\end{align}
After fixing the slack bus voltage to unity, the environment then solves the network equations given by (\ref{eq:power_flow_equations}). Our implementation uses the Newton-Raphson procedure \cite{sun1984optimal} to do so. From the solution, we obtain the voltage $V_i^{(bus)}$ at each non-slack bus and the slack generator power injection point $(P_{g^{slack},t+1}, Q_{g^{slack},t+1})$.

\paragraph{5. State construction} The new state vector $s_{t+1}$ can now be constructed according to the structure defined by (\ref{eq:state}). The active and reactive power injection points $P_{d,t+1}$, $Q_{d,t+1}$ have already been computed. The new charge level $SoC_{d,t+1}$ of each DES unit $d \in \mathcal D_{DES}$ is obtained using expression (\ref{eq:next_soc}). Finally, the $P^{(max)}_{g,t+1}$ and $aux_{t+1}^{(k)}$ variables are simply copied from the output of \lstinline{next_vars()}.

\subsection{Reward Function}
\label{app:reward_function}

The main component of the reward signal, as introduced in (\ref{eq:full-reward}) and (\ref{eq:reward-ct}), is a sum of three energy losses and a penalty term associated with violating operating constraints:
\begin{align}
    c_t = -\big(\Delta E_{t:t+1} + \lambda \phi(s_{t+1}) \big).
\end{align}
We chose to compute both terms in p.u. to ensure similar orders of magnitude.

\subsubsection{Energy loss}
The transmission energy loss, $\Delta E^{(1)}_{t:t+1}$, is computed as:
\begin{align}
    \Delta E^{(1)}_{t:t+1} = \Delta t \sum_{d \in \mathcal D} P_{d,t+1}^{(dev)}, \label{eq:E1}
\end{align}
where $\Delta t$ is used to get the energy loss in p.u. per hour. The net amount of energy flowing from the grid into DES units, $\Delta E^{(2)}_{t:t+1}$, is obtained using:
\begin{align}
    \Delta E^{(2)}_{t:t+1} = - \Delta t \sum_{d \in \mathcal D_{DES}} P_{d,t+1}^{(dev)}. \label{eq:E2}
\end{align}
Finally, the amount of energy loss as a result of renewable energy curtailment, $\Delta E^{(3)}_{t:t+1}$, is:
\begin{align}
    \Delta E^{(3)}_{t:t+1} =  \Delta t \sum_{g \in  \mathcal D_{RER}} (P_{g,t+1}^{(max)} - P^{(dev)}_{g,t+1}). \label{eq:E3}
\end{align}
Summing (\ref{eq:E1})-(\ref{eq:E3}) together yields the total energy loss:
\begin{align}
    \Delta E_{t:t+1} = \Delta t \Big( \sum_{d \in \mathcal D_G \cup \mathcal D_L} P^{(dev)}_{d,t+1} + \sum_{g \in  \mathcal D_{RER}} (P_{g,t+1}^{(max)} - P^{(dev)}_{g,t+1}) \Big) .
\end{align}

\subsubsection{Constraint-violation penalty}
Let $\Phi: \mathcal S \to \mathbb R$ be the penalty function that adds a large cost $\lambda \Phi(s_{t+1})$ to a policy that leads to a violation of operating constraints. To compute $\Phi(s_{t+1})$, the environment first computes the node voltages $V_{i,t+1}$ using (\ref{eq:power_flow_equations}) and the directed branch currents $I_{ij,t+1}$ and $I_{ji,t+1}$ for each branch $e_{ij} \in \mathcal E$ using (\ref{eq:branch_currents}). The obtained values are then plugged into $|S_{ij,t+1}| = |V_{i,t+1}I_{ij,t+1}^*|$ and $|S_{ji,t+1}| =  |V_{j,t+1} I_{ji,t+1}^*|$ to compute the corresponding branch's apparent power flows. The penalty term $\Phi(s_{t+1})$ is finally obtained using:
\begin{align}
\Phi(\mathbf s_{t+1}) = \Delta t \Big(&\sum_{i \in \mathcal N} \big(\max{(0, |V_{i,t+1}| - \overline V_i)} + \max{(0, \underline V_i - |V_{i,t+1}|)} \big) \nonumber \\
+ &\sum_{e_{ij} \in \mathcal E} \max{(0, |S_{ij,t+1}| - \overline S_{ij}, |S_{ji,t+1}| - \overline S_{ij})} \Big) \;.
\end{align}

\clearpage
\section{Model Predictive Control Scheme}
\label{app:opf}

\subsection{Introduction}
This appendix describes the MPC problem solved by the MPC-based policy $\pi_{MPC-N}$ introduced in Section \ref{sec:baseline}. At each timestep, the policy solves a multi-stage DCOPF problem with an optimization horizon of $N$ timesteps. As a linear approximation of the actual ACOPF that we would like to solve, the DCOPF formulation relies on three assumptions, included here again for clarity:
\begin{enumerate}
    \item Transmission lines are lossless: $r_{ij} = 0, \; \forall e_{ij} \in \mathcal E$,
    \item The difference between adjacent bus voltage angles is small: $\angle V_i \approx \angle V_j$, $\forall e_{ij} \in \mathcal E$,
    \item Bus voltage magnitudes are close to unity: $|V_i| \approx 1, \; \forall i \in \mathcal N$.
\end{enumerate}

We start by giving a general formulation of the MPC problem in which the algorithm takes as input predictions of future demand and generation in Section \ref{app:mpc-general}. We call this policy $\pi_{MPC-N}$. We then consider two particular forecasting methods in Section \ref{app:mpc-special-cases}: one which assumes constant values over the optimization horizon, policy $\pi_{MPC-N}^{constant}$, and another that generates perfect forecasts, policy $\pi_{MPC-N}^{perfect}$.

\subsection{General formulation}
\label{app:mpc-general}
\subsubsection{Policy overview}
The action selection procedure followed by $\pi_{MPC-N}$ at timestep $t$ is given by Algorithm \ref{algo:baseline}. In this algorithm, \lstinline{solveMPC()} refers to solving\footnote{Our implementation uses the CVXPY Python optimization package \cite{diamond2016cvxpy, agrawal2018rewriting} to solve the optimization program.} the optimization problem (\ref{eq:optim-obj})-(\ref{eq:optim-slack-angle}) and extracting the vector of device active power injection $\mathbf P_{t+1}^{(dev)}$ from the solution. The considered-optimal power injections from all non-slack generators and DES units are then concatenated into an action vector $a_t$. Since reactive power flows are ignored by the DCOPF formulation, we chose to simply set the reactive power set-points in $a_t$ to zero.

In this general formulation, Algorithm \ref{algo:baseline} takes as inputs the network state $s_t$ (directly extracted from the Gym-ANM simulator) and forecasts of demand and generation over the optimization horizon $k =t+1,\ldots, t+N$. We denote these forecasted values as $\tilde P^{(dev)}_{l,k}$ and $\tilde P_{g,k}^{(max)}$, respectively. An additional safety margin hyperparameter, $\beta \in [0,1]$, is also introduced to further constrain the power flow on each transmission line in the OPF. This is done with the hope that it will account for any errors introduced with the linear DC approximation, thus ensuring that line current constraints are respected. The penalty hyperparameter $\lambda$ is taken to be the same as in the reward function.

\begin{algorithm}
\setstretch{1.3}
\caption{MPC (multi-stage DCOPF) policy $\pi_{MPC-N}$.}
\label{algo:baseline}
\begin{algorithmic}[1]
\STATE \textbf{Input:} State $s_t$, demand forecasts $\{\tilde P_{l,t+k}^{(dev)}\}_{l\in \mathcal D_L, k=1,\ldots,N}$, maximum generation forecasts $\{\tilde P_{g,t+k}^{(max)}\}_{g\in \mathcal D_G - \{g^{slack}\}, k=1,\ldots,N}$
\STATE \textbf{Parameter:} Safety margin $\beta \in [0,1]$, penalty hypeparameter $\lambda$
\STATE $\{P_{d,t+1}^{(dev)}\}_{d \in \mathcal D} \gets$ \lstinline{solveMPC(}$s_t, \{\tilde P_{l,t+k}^{(dev)}\}, \{\tilde P_{g,t+k}^{(max)}\}, \beta,\lambda, \text{grid\_characteristics}$\lstinline{)}
\FOR{$g \in \mathcal D_G - \{g^{slack}\}$}
    \STATE $a_{P_{g,t}} \gets P_{g,t+1}^{(dev)}$
    \STATE $a_{Q_{g,t}} \gets 0$
\ENDFOR
\FOR{$d \in \mathcal D_{DES}$}
    \STATE $a_{P_{d,t}} \gets P_{d,t+1}^{(dev)}$
    \STATE $a_{Q_{d,t}} \gets 0$
\ENDFOR
\end{algorithmic}
\end{algorithm}

\subsubsection{The optimization problem}
We now describe the optimization problem (\ref{eq:optim-obj})-(\ref{eq:optim-slack-angle}) in more detail. The objective function is a simplified version of the cost function used in the reward signal, originally defined as:
\begin{align}
    \sum_{g \in \mathcal D_G \cup \mathcal D_L} P_{d,t+1}^{(dev)} + \sum_{g \in \mathcal D_{RER}} \big( P_{g,t+1}^{(max)} - P_{g,t+1}^{(dev)} \big) + \lambda \phi(s_{t+1}) .
\end{align}
In the above formulation, load injections and maximum generations of generators are non-controllable variables (i.e., constants), which can thus be removed from the objective function. In addition, the DCOPF assumptions have $|V_i| = 1$, which leads to $|S_{ij}|=|P_{ij}|$, from which the penalty term $\phi(s_{t+1})$ can be greatly simplified. The resulting objective function to minimize is given by (\ref{eq:optim-obj}). Note that we define it as the discounted sum of costs over the optimization horizon. This is to reflect the agent's objective of learning a policy that minimizes the expected discounted return.

Constraints (\ref{eq:optim-bus-dev-relation}) and (\ref{eq:optim-bus-angle-relation}) express the relationships between nodal power injections, device power injections, and bus voltage angles. Branch power flow equations are formalized by (\ref{eq:optim-power-flows}). Equalities (\ref{eq:optim-load-injections}) constrain the load power injections in the vector $\mathbf P_k^{(dev)}$ to the specified forecasted values. Similarly, expression (\ref{eq:optim-max-generation}) uses the forecasted generation upper bounds to limit generator injections in $\mathbf P_k^{(dev)}$. Both DES devices and non-slack generators are restricted to their physical range of operation in (\ref{eq:optim-gen-des-bounds}), assuming reactive power injections of zero. In (\ref{eq:optim-soc-bounds}), power injections from DES units are limited to values that ensure $SoC_{d,k+1} \in [\underline {SoC}_d, \overline {SoC}_d]$. Finally, (\ref{eq:optim-angle-bounds}) constrains voltage angles to be within $[0, 2\pi]$ radians and (\ref{eq:optim-slack-angle}) provides a voltage angle reference by fixing the slack voltage angle to 0.

{
\begin{align}
    &\underset{\substack{\mathbf P_{k}^{(dev)}, \mathbf V_{k}, \\ k=t+1,\ldots,t+N}}{\text{minimize}}&& \sum_{k=t+1}^{t+N} \gamma^{k-t-1} \Big( \sum_{g \in \mathcal D_G - \mathcal D_{RER}} P_{g,k}^{(dev)} + \lambda \sum_{e_{ij} \in \mathcal E} \max{(0, |P_{ij,k}| - \beta \overline S_{ij})} \Big) \label{eq:optim-obj} \\
    &\text{subject to}&& P_{i, k}^{(bus)} = \sum_{d \in \mathcal D_i} P_{d, k}^{(dev)}, \quad \forall i \in \mathcal N, \forall k  \label{eq:optim-bus-dev-relation} \\
    &&& P^{(bus)}_{i,k} = \sum_{e_{ij} \in \mathcal E} B_{ij}(\angle V_{i,k} - \angle V_{j,k}) + \sum_{e_{ji} \in \mathcal E} B_{ji} (\angle V_{i, k} - \angle V_{j, k}),  \forall i \in \mathcal N, \forall k \label{eq:optim-bus-angle-relation}\\
    &&& P_{ij,k} = B_{ij} (\angle V_{i,k} - \angle V_{j,k}), \quad \forall e_{ij} \in \mathcal E, \forall k \label{eq:optim-power-flows} \\
    &&& P_{l,k}^{(dev)} = \tilde P_{l,k}^{(dev)}, \quad \forall l \in \mathcal D_L, \forall k \label{eq:optim-load-injections} \\
    &&& P_{g,k}^{(dev)} \le \tilde P_{g,k}^{(max)}, \quad \forall g \in \mathcal D_{G} - \{g^{slack}\}, \forall k \label{eq:optim-max-generation} \\
    &&& \underline P_d \le P^{(dev)}_{d,k} \le \overline P_{d}, \quad \forall d \in \mathcal D_G \cup \mathcal D_{DES} - \{g^{slack}\}, \forall k \label{eq:optim-gen-des-bounds} \\
    &&& \frac{1}{\Delta t \eta}(SoC_{d,k} - \overline{SoC}_d) \le P_{d,k}^{(dev)} \le \frac{\eta}{\Delta t}(SoC_{d,k} - \underline{SoC}_d), \forall d \in \mathcal D_{DES}, \forall k \label{eq:optim-soc-bounds} \\
    &&& 0 \le \angle V_{i,k} \le 2\pi, \quad \forall i \in \mathcal N, \forall k \label{eq:optim-angle-bounds} \\
    &&& \angle V_{0,k} = 0, \quad \forall k \label{eq:optim-slack-angle}
\end{align}
}

\subsubsection{Further considerations}
The performance achieved by $\pi_{MPC-N}$ provides a lower bound on the best performance achievable in a given environment. This bound is not tight, however, since the achieved performance depends on (a) the quality of the DC linear approximation, (b) the accuracy of the forecasted values, and (c) the length of the optimization horizon $N$. Note that, in general, the performance of an MPC-based policy increases as $N \to \infty$. Because of (a) and (b), however, this may not be the case with $\pi_{MPC-N}$, since, e.g., erroneous long-term forecasts may harm policies with larger $N$'s. As a result, $N$ may have to be tuned, depending on the environment.

\subsection{Special cases: constant and perfect forecast}
\label{app:mpc-special-cases}
We now consider two special cases of the MPC-based policy $\pi_{MPC-N}$. Both policies were used in the ANM6-Easy environment in Section \ref{sec:experiments}.

\subsubsection{Constant forecast}
The first variant that we consider is $\pi_{MPC-N}^{constant}$. It assumes that load injections $P_{l,t}^{(dev)}$ and maximum generations $P_{g,t}^{(max)}$ remain constant during the optimization horizon. As such, it is one of the simplest variants of $\pi_{MPC-N}$ that one could use. Formally, we can describe $\pi_{MPC-N}^{constant}$ by its constant forecasts:
\begin{align}
\begin{cases}
    \tilde P_{l,k}^{(dev)} = P_{l,t}^{(dev)}, & \forall l \in \mathcal D_L,\; k=t+1,\ldots,t+N, \\
    \tilde P_{g,k}^{(max)} = P_{g,t}^{(max)}, & \forall g \in \mathcal D_G - \{g^{slack}\},\; k=t+1,\ldots,t+N .
\end{cases}
\end{align}

The main advantage of $\pi_{MPC-N}^{constant}$ is that it can be used out-of-the-box in any Gym-ANM environment. More information on how to do this can be found on the project repository.

\subsubsection{Perfect forecast}
The second variant that we consider is $\pi_{MPC-N}^{perfect}$. This variant is specifically tailored for the ANM6-Easy environment introduced in Section \ref{sec:anm6-easy}. This is because it assumes perfect forecasts of load injections and maximum generations. In other words, it relies on the fact that ANM6-Easy is a deterministic environment in which future demand and generation can be perfectly predicted. Formally, $\pi_{MPC-N}^{perfect}$ uses perfect forecasts:
\begin{align}
    \begin{cases}
    \tilde P_{l,k}^{(dev)} = P_{l,k}^{(dev)}, & \forall l \in \mathcal D_L,\; k=t+1,\ldots,t+N, \\
    \tilde P_{g,k}^{(max)} = P_{g,k}^{(max)}, & \forall g \in \mathcal D_G - \{g^{slack}\},\; k=t+1,\ldots,t+N .
\end{cases}
\end{align}

Unlike $\pi_{MPC-N}^{constant}$, policy $\pi_{MPC-N}^{perfect}$ can only be used in deterministic environments of the like of ANM6-Easy. Nevertheless, it offers a large advantage in that it yields a much better performance in such environments. This provides the user with a tighter lower bound on the best achievable performance in the environment.

\clearpage
\section{New Gym-ANM Environments}
\label{app:new_env}

This appendix gives an overview of the procedure to follow to design new Gym-ANM environments\footnote{Further guidelines and tutorials can be found on the project repository.}. New Gym-ANM environments can be implemented as Python subclasses that inherit the provided \lstinline{ANMEnv} superclass, following the template presented in Listing \ref{lst:new_env}.

{\footnotesize
\begin{lstlisting}[language=Python, caption={Implementation template for new Gym-ANM environments.}, label=lst:new_env, captionpos=b, frame=single, escapeinside={(*}{*)}, xleftmargin=3.4pt, xrightmargin=3.4pt]
from gym_anm.envs import ANMEnv

class MyANMEnv(ANMEnv):
    def __init__(self):
        network = {'baseMVA':..., 'bus':..., 'device':..., 
                   'branch':...}
        obs = [('bus_p', [0,1], 'MW'), ('dev_q', [2], 'MW')] 
              # or a callable
        K = 1
        delta = 0.25
        gamma = 0.999
        lamb = 1000     # 'lambda' is a reserved keyword in Python
        r_clip = 100
        seed = None
        super().__init__(network, obs, K, delta, gamma, lamb, 
                         r_clip, seed)
    def init_state(self):
        ...
    def next_vars(self, s_t):
        ...
    def observation_bounds(self):  # optional
        ...
\end{lstlisting}}

\paragraph{\lstinline{__init__()}}
This method, known as a constructor in object-oriented languages, is called when a new instance of the new environment \lstinline{MyANMEnv} is created. In order to initialize the environment, the following arguments need to be passed to the superclass, through the call \lstinline{super().__init__()}:
\begin{itemize}
    \item \lstinline{network}: a Python dictionary that describes the structure and characteristics of the distribution network $G$ and the set of electrical devices $\mathcal D$. Its structure should follow the one given in Appendix \ref{app:case_file}. 
    \item \lstinline{obs}: a list of tuples corresponding to the variables to include in observation vectors. In Listing \ref{lst:new_env}, $o_t$ is constructed as $(P^{(bus)}_{0,t}, P^{(bus)}_{1,t}, Q^{(dev)}_{2,t})$, all in MW units. The full list of supported combinations is given in Table \ref{tab:obs_available_values}. Alternatively, the \lstinline{obs} object can be defined as a customized callable object (function) that returns observation vectors when called (i.e., $o_t =$ \lstinline{obs(}$s_t$\lstinline{)}), or as a string \lstinline{'state'}. In the later case, the environment becomes fully observable and observations $o_t = s_t$ are emitted.
    \item \lstinline{K}: the number of auxiliary variables $K$ in the state vector given by (\ref{eq:state}).
    \item \lstinline{delta}, \lstinline{gamma}, \lstinline{lamb}, \lstinline{r_clip}: the hyperparameters $\Delta t$, $\gamma$, $\lambda$, and $r_{clip}$, respectively, used to compute the rewards and returns, as introduced in Section \ref{sec:gym_anm_overview}. 
    \item \lstinline{seed}: an integer to be used as random seed.
\end{itemize}

\paragraph{\lstinline{init_state()}} 
This method will be called once at the start of each new trajectory and it should return an initial state vector $s_0$ that matches the structure of (\ref{eq:state}). In the case where $s_0$ falls outside of $\mathcal S$ because, for instance, the $(P, Q)$ injection point of a device falls outside of its operating range, the environment will map $s_0$ to the closest element of $\mathcal S$ according to the Euclidean distance. In short, \lstinline{init_state()} implements $p_0(\cdot)$.

\paragraph{\lstinline{next_vars()}} As introduced in Section \ref{sec:gym_anm_overview}, \lstinline{next_vars()} is a method that receives the current state vector $s_t$ and should return the outcomes of the internal variables for timestep $t+1$. It must be implemented by the designer of the task, with the only constraint being that it must return a list of $|\mathcal D_L| + |\mathcal D_{RER}| + K$ values.

\paragraph{\lstinline{observation_bounds()}} This method is optional and only useful if the observation space is specified as a callable object. In the latter case, \lstinline{observation_space()} should return the (potentially loose) bounds of the observation space $\mathcal O$, so that agents can easily normalize emitted observation vectors. \\

Additional \lstinline{render()} and \lstinline{close()} methods can also be implemented to support rendering of the interactions between the agent and the new environment. \lstinline{render()} should update the visualization every time it gets called, and \lstinline{close()} should end the rendering process. For more information, we refer to the official OpenAI Gym documentation \cite{brockman2016openai}.\footnote{\url{https://gym.openai.com/docs/}}

\begin{table}[]
\centering
\begin{tabular}{lll}
 \toprule
 Keyword & Description & Units \\
 \midrule
bus\_p (dev\_p) & Bus (Device) active power injection $P^{(bus)}_i$ ($P^{(dev)}_d$) & MW, pu \\
bus\_q (dev\_q) & Bus (Device) reactive power injection $Q^{(bus)}_i$ ($Q^{(dev)}_d$) & MVAr, pu \\
bus\_v\_magn & Bus voltage magnitude $|V_i|$ & pu, kV \\
bus\_v\_ang & Bus voltage angle $\angle V_i$ & degree, rad \\
bus\_i\_magn & Bus current injection magnitude $|I_i|$ & pu, kA \\
bus\_i\_ang & Bus current injection angle $\angle I_i$ & degree, rad \\
branch\_p & Branch active power flow $P_{ij}$ & MW, pu \\
branch\_q & Branch reactive power flow $Q_{ij}$ & MVAr, pu \\
branch\_s & Branch apparent power flow $|S_{ij}|$ & MVA, pu \\
branch\_i\_magn & Branch current magnitude $|I_{ij}|$ & pu \\
branch\_i\_ang & Branch current angle $\angle I_{ij}$ & degree, rad \\
des\_soc & SOC of DES $SoC_{d}$ & MWh, pu \\
gen\_p\_max & Generator dynamic upper bound $P_g^{(max)}$ & MW, pu \\
aux & Vector of $K$ auxiliary variables $aux^{(K)}$ & -- \\
 \bottomrule
\end{tabular}
\vspace{0.2cm}
\caption{Available combinations for the \lstinline{observation} parameter. Each type of observation should be provided as a tuple with the corresponding bus/device indices or with the \lstinline{'all'} keyword. Units can also be specified. For instance: \lstinline{[('bus_p', 'all', 'pu'), ('dev_q', [1,2], 'MVAr'), ('branch_s', [(1,2)])]} would lead to observation vectors $o_t = [P_1^{(bus)}, \ldots, P_N^{(bus)}, Q_1^{(dev)}, Q_2^{(dev)}, |S_{12}|]$.}
\label{tab:obs_available_values}
\end{table}

\clearpage
\section{Network Input Dictionary}
\label{app:case_file}

This appendix describes the structure of the Python dictionary required to build new Gym-ANM environments. The dictionary should contain four keys: \lstinline{'baseMVA'}, \lstinline{'bus'}, \lstinline{'device'}, and \lstinline{'branch'}. The value given to the key \lstinline{'baseMVA'} should be a single integer, representing the base power of the system (in MVA) used to normalize values to per-unit. Each of the other three keys should be associated with a numpy 2D array, in which each row represents a single bus, device, or branch of the distribution network. The structures of the \lstinline{'bus'}, \lstinline{'device'}, and \lstinline{'branch'} arrays are described in Tables \ref{tab:network_file_bus}, \ref{tab:network_file_device}, and \ref{tab:network_file_branch}, respectively.

\begin{table}[h]
\centering
\begin{tabular}{ll}
 \toprule
 Column & Description \\
 \midrule
 0 & Bus unique ID $i$ (0-indexing). \\
 1 & Bus type (0 = slack, 1 = PQ). \\
 2 & RMS base voltage of the zone (kV). \\
 3 & Maximum RMS voltage magnitude $\overline V_i$ (p.u.). \\
 4 & Minimum RMS voltage magnitude $\underline V_i$ (p.u.). \\
 \bottomrule
\end{tabular}
\vspace{0.2cm}
\caption{Bus data: description of each row in the \lstinline{'bus'} numpy array.}
\label{tab:network_file_bus}
\end{table}

\begin{table}[h]
\centering

 \begin{tabular}{p{.09\columnwidth}p{0.8\textwidth}} 
 \toprule
 Column & Description \\
 \midrule
 0 & Device unique ID $d$ (0-indexing). \\
 1 & Bus unique ID $i$ to which $d$ is connected. \\
 2 & Type of device (-1 = load; 0 = slack; 1 = classical generator; 2 = distributed renewable energy generator; 3 = DES unit). \\
 3 & Constant ratio of reactive power over active power $(Q/P)_d$ (loads only). \\ 
 4 & Maximum active power output $\overline P_d$ (MW). \\
 5 & Minimum active power output $\underline P_d$ (MW). \\
 6 & Maximum reactive power output $\overline Q_d$ (MVAr). \\
 7 & Minimum reactive power output $\underline Q_d$ (MVAr). \\
 8 & Positive active power output of PQ capability curve $P^+_d$ (MW). \\
 9 & Negative active power output of PQ capability curve $P^-_d$ (MW). \\
 10 & Positive reactive power output of PQ capability curve $Q^+_d$ (MVAr). \\
 11 & Negative reactive power output of PQ capability curve $Q^-_d$ (MVAr). \\
 12 & Maximum state of charge of storage unit $\overline{SoC}_d$ (MWh). \\
 13 & Minimum state of charge of storage unit $\underline{SoC}_d$ (MWh). \\
 14 & Charging and discharging efficiency coefficient of storage unit $\eta_d$. \\
 \bottomrule
\end{tabular}
\vspace{0.2cm}
\caption{Device data: description of each row in the \lstinline{'device'} numpy array.}
\label{tab:network_file_device}
\end{table}

\begin{table}[h]
\centering
 \begin{tabular}{p{.09\columnwidth}p{0.8\textwidth}} 
 \toprule
 Column & Description \\
 \midrule
0 & Sending-end bus unique ID $i$. \\
1 & Receiving-end bus unique ID $j$. \\
2 & Branch series resistance $r_{ij}$ (p.u.). \\
3 & Branch series reactance $x_{ij}$ (p.u.). \\
4 & Branch total charging susceptance $b_{ij}$ (p.u.). \\
5 & Branch rating $\overline S_{ij}$ (MVA). \\
6 & Transformer off-nominal turns ratio $\tau_{ij}$. \\
7 & Transformer phase shift angle $\theta_{ij}$ (degrees) ($>0$ = delay). \\
 \bottomrule
\end{tabular}
\vspace{0.2cm}
\caption{Branch data: description of each row in the \lstinline{'branch'} numpy array.}
\label{tab:network_file_branch}
\end{table}

\clearpage
\section{ANM6-Easy Environment}
\label{app:smartgrid-env6_specs}

This appendix describes in more detail the ANM6-Easy Gym-ANM environment introduced in Section \ref{sec:anm6-easy}. 

\subsection{Network Characteristics}
\begin{figure}[h]
\begin{minipage}{.5\linewidth}
\centering
\subfloat[]{\label{fig:inj_points_solar}\includegraphics[scale=.45]{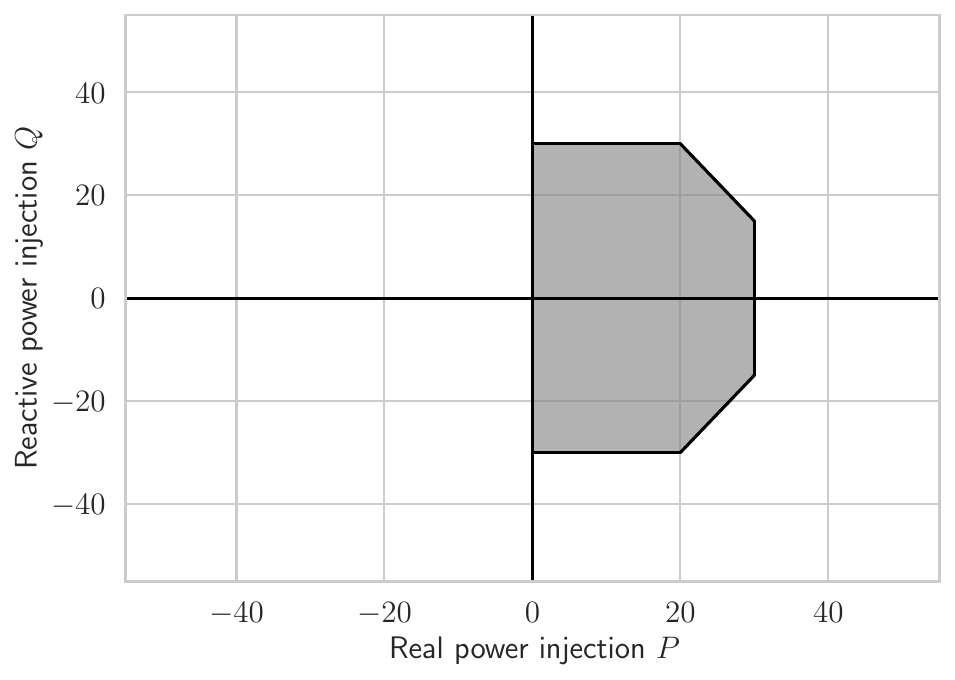}}
\end{minipage}%
\begin{minipage}{.5\linewidth}
\centering
\subfloat[]{\label{fig:inj_points_wind}\includegraphics[scale=.45]{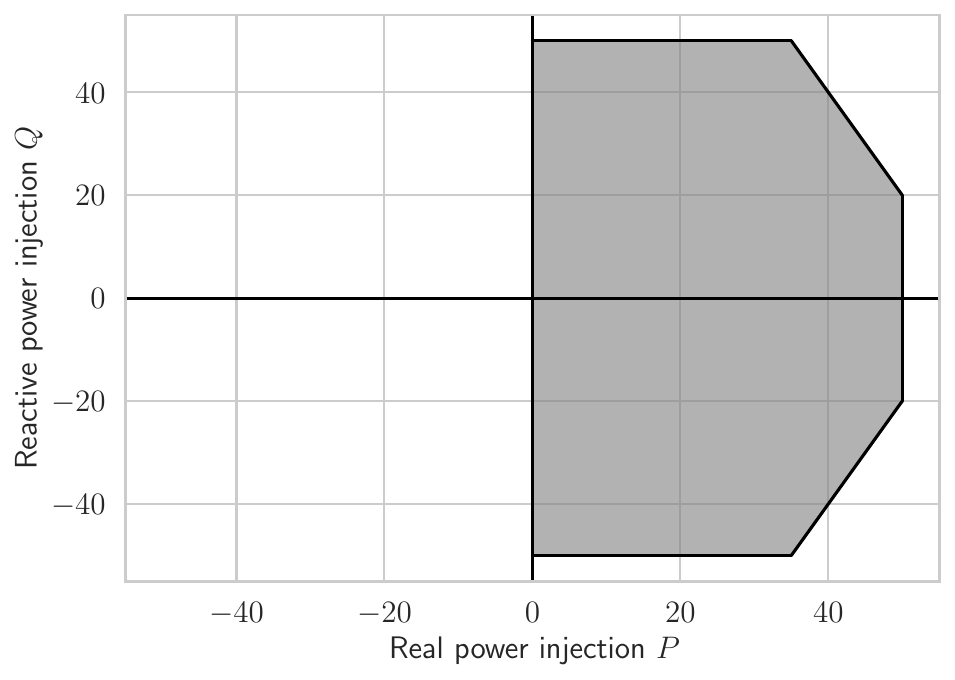}}
\end{minipage}\par\medskip
\centering
\subfloat[]{\label{fig:inj_points_des}\includegraphics[scale=.45]{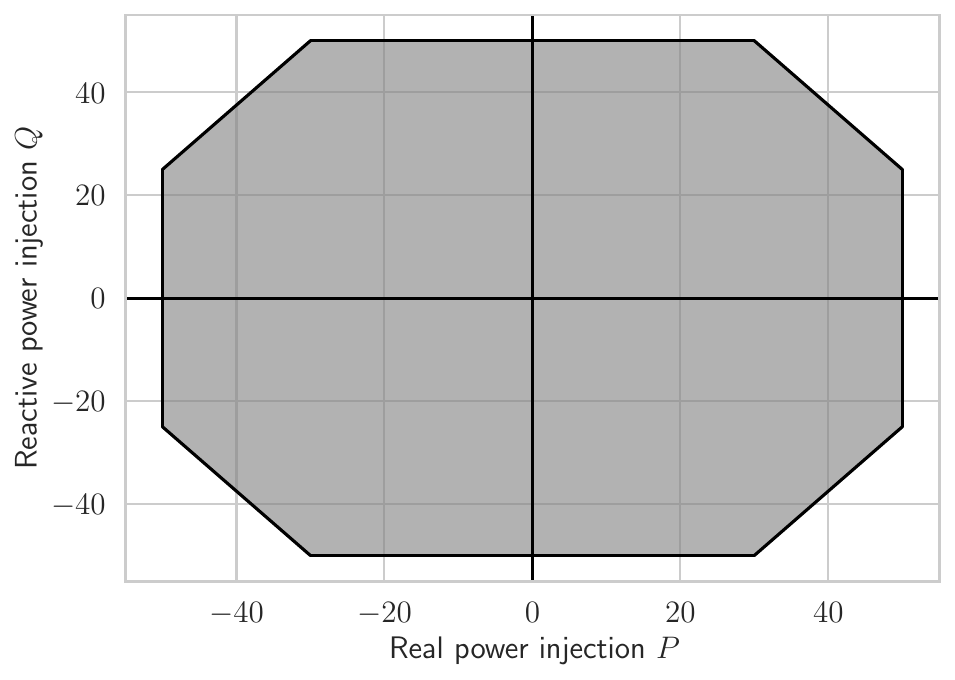}}
\caption{The range of valid $(P, Q)$ injection points for (a) the solar farm, (b) the wind farm, and (c) the DES unit of ANM6-Easy, as formalized in Appendices \ref{app:distributed_generators} and \ref{app:des}.}
\label{fig:anm6_inj_points}
\end{figure}

Tables \ref{tab:bus_specs}, \ref{tab:device_specs}, and \ref{tab:branch_specs} summarize the characteristics of buses, electrical devices, and branches, respectively, following the network input dictionary structure given in Appendix \ref{app:case_file}. Based on the values given in Table \ref{tab:device_specs}, the range of operation (see Appendices \ref{app:distributed_generators} and \ref{app:des}) of the distributed generators and the DES unit of ANM6-Easy are also plotted in Figure \ref{fig:anm6_inj_points}.

\begin{table}[h]\centering
    \ra{1.2}
    \begin{tabular}[t]{ccccc}\toprule
        $i$ & Type & Base voltage & $\overline V_i$ & $\underline V_i$ \\
        \midrule
         0 & 0 & 132 & 1.04 & 1.04 \\
         1 & 1 & 33  & 1.1  & 0.9  \\
         2 & 1 & 33  & 1.1  & 0.9 \\
         3 & 1 & 33  & 1.1  & 0.9 \\
         4 & 1 & 33  & 1.1  & 0.9 \\
         5 & 1 & 33  & 1.1  & 0.9 \\
        \bottomrule
    \end{tabular}
    \vspace{0.2cm}
    \caption{Description of each bus $i \in \mathcal N$ of ANM6-Easy.}
    \label{tab:bus_specs}
\end{table}
\begin{table}[h]\centering
\ra{1.2}
    \begin{tabular}{c@{\hspace{8pt}}c@{\hspace{8pt}}c@{\hspace{8pt}}c@{\hspace{8pt}}c@{\hspace{8pt}}c@{\hspace{8pt}}c@{\hspace{8pt}}c@{\hspace{8pt}}c@{\hspace{8pt}}c@{\hspace{8pt}}c@{\hspace{8pt}}c@{\hspace{8pt}}c@{\hspace{8pt}}c@{\hspace{8pt}}c@{\hspace{8pt}}}\toprule
        $d$ & $i$ & Type & $(Q/P)_d$ & $\overline P_d$ & $\underline P_d$ & $\overline Q_d$ & $\underline Q_d$ & $P^+_d$ & $P^-_d$ & $Q^+_d$ & $Q^-_d$ & $\overline{SoC}_d$ & $\underline{SoC}_d$ & $\eta_d$ \\
        \midrule
 0 & 0 & 0 &  -  & -  & -   & -  & -   & -  & -   & -  & -   & -   & - & -   \\
 1 & 3 & -1 & 0.2 & 0  & -10 & -  & -   & -  & -   & -  & -   & -   & - & -   \\
 2 & 3 & 2 &  -  & 30 & 0   & 30 & -30 & 20 & -   & 15 & -15 & -   & - & -   \\
 3 & 4 & -1 & 0.2 & 0  & -30 & -  & -   & -  & -   & -  & -   & -   & - & -   \\
 4 & 4 & 2 &  -  & 50 & 0   & 50 & -50 & 35 & -   & 20 & -20 & -   & - & -   \\
 5 & 5 & -1 & 0.2 & 0  & -30 & - & -    & -  & -   & -  & -   & -   & - & -   \\
 6 & 5 & 3 &  -  & 50 & -50 & 50 & -50 & 30 & -30 & 25 & -25 & 100 & 0 & 0.9 \\
 \bottomrule
    \end{tabular}
    \vspace{0.2cm}
    \caption{Description of each electrical device $d \in \mathcal D$ of ANM6-Easy.}
    \label{tab:device_specs}
\end{table}
\begin{table}[h]
\centering
 \begin{tabular}[h]{llcccccc} \toprule
 i & j & $r_{ij}$ & $x_{ij}$ & $b_{ij}$ & $\overline S_{ij}$ & $\tau_{ij}$ & $\theta_{ij}$ \\
 \midrule
0 & 1 & 0.0036 & 0.1834 & 0 & 32 & 1 & 0 \\
1 & 2 & 0.03   & 0.022  & 0 & 25 & 1 & 0 \\
1 & 3 & 0.0307 & 0.0621 & 0 & 18 & 1 & 0 \\
2 & 4 & 0.0303 & 0.0611 & 0 & 18 & 1 & 0 \\
2 & 5 & 0.0159 & 0.0502 & 0 & 18 & 1 & 0 \\
 \bottomrule
\end{tabular}
\vspace{0.2cm}
\caption{Description of each branch $e_{ij} \in \mathcal E$ of ANM6-Easy.}
\label{tab:branch_specs}
\end{table}

The fixed time series used by the \lstinline{next_vars()} component of ANM6-Easy, in order to model the evolution of the loads and of the maximum generation from renewable energy resources, are provided in Table \ref{tab:time_series}. 

\begin{table}[h]
\small 
\centering
 \begin{tabular}[h]{llllllll} \toprule
 Situation & \multicolumn{2}{c}{$aux_t^{(0)}$} & $\mathbf P_1$ & $\mathbf P_2$ & $\mathbf P_3$ & $\mathbf P_4$ & $\mathbf P_5$ \\
 \midrule
 1 & 0-24 & 92-95 & -1 & 0 & -4 & 40 & 0 \\
 1-2 & 25 & 91 & -1.5 & 0.5 & -4.75 & 36.375 & -3.125 \\
 1-2 & 26 & 90 & -2 & 1 & -5.5 & 32.75 & -6.25 \\
 1-2 & 27 & 89 & -2.5 & 1.5 & -6.25 & 29.125 & -9.375 \\
 1-2 & 28 & 88 & -3 & 2 & -7 & 25.5 & -12.5 \\
 1-2 & 29 & 87 & -3.5 & 2.5 & -7.75 & 21.875 & -15.625 \\
 1-2 & 30 & 86 & -4 & 3 & -8.5 & 18.25 & -18.75 \\
 1-2 & 31 & 85 & -4.5 & 3.5 & -9.25 & 14.625 & -21.875 \\
 2 & 32-44 & 72-84 & -5 & 4 & -10 & 11 & -25 \\
 2-3 & 45 & 71 & -4.625 & 7.25 & -11.25 & 14.625 & -21.875 \\
 2-3 & 46 & 70 & -4.25 & 10.50 & -12.5 & 18.25 & -18.75 \\
 2-3 & 47 & 69 & -3.875 & 13.75 & -13.75 & 21.875 & -15.625 \\
 2-3 & 48 & 68 & -3.5 & 17 & -15 & 25.5 & -12.5 \\
 2-3 & 49 & 67 & -3.125 & 20.25 & -16.25 & 29.125 & -9.375 \\
 2-3 & 50 & 66 & -2.75 & 23.5 & -17.5 & 32.75 & -6.25 \\
 2-3 & 51 & 65 & -2.375 & 26.75 & -18.75 & 36.375 & -3.125 \\
 3 & 52-64 & & -2 & 30 & -20 & 40 & 0 \\
 \bottomrule
\end{tabular}
\vspace{0.2cm}
\caption{The fixed time series $\mathbf P_{1-5}$ used to model the temporal evolution of the loads $P_{l,t}^{(dev)}$, $l \in \{1,3,5\}$, and the maximum generations $P_{g,t}^{(max)}$, $g \in \{2,4\}$, in the ANM6-Easy environment, creating the three challenging situations described in Section \ref{sec:environments}. The auxiliary variable $aux_t^{(0)} = (t_0 + t) \mod{\frac{24}{\Delta t}}$ is used as an index to those time series.}
\label{tab:time_series}
\end{table}

\subsection{Environment Initialization}
The initialization procedure of ANM6-Easy, according to which initial states $s_0 \sim p_0(\cdot)$ are drawn, is illustrated in Algorithm \ref{algo:anm6-easy}. Time series $\mathbf P_{1-5}$ refer to Table \ref{tab:time_series}. An initial time of day $t_0$ is sampled and used to initialize the $aux^{(0)}$ variable (lines 2-3) and to index the fixed time series of active power demand and maximum generation (lines 5, 8). In line 5, the $(P,Q)$ power injection from each load is obtained based on their respective constant power factor. We assume that each distributed generator operates at its maximum active power (i.e., no generator is curtailed) and that its reactive power injection is sampled uniformly. The initial power injection point of each generator is then mapped onto the generator's allowed region of operation $\mathcal R_{g,0}$ (line 9). The initial state of charge of the DES unit is also uniformly sampled and its power injection point is set to zero (lines 12-13). Finally, the slack power injection is obtained after solving the set of network equations (line 15).

\begin{algorithm}
\setstretch{1.3}
\caption{Initialization of ANM6-Easy, $p_0(\cdot)$.}
\label{algo:anm6-easy}
\begin{algorithmic}[1]
\STATE \textbf{Output:} $s_0 \in \mathcal S$
\STATE $t_0 \sim U\{0,\frac{24}{\Delta t}-1\}$
\STATE $aux_0^{(0)} \gets t_0$
\FOR{$l \in \mathcal D_L$}
\STATE $(P_{l,0}^{(dev)}, Q_{l,0}^{(dev)})  \gets (\mathbf{P}_l[t_0], \mathbf{P}_l[t_0] \tan{\phi_l})$
\ENDFOR
\FOR{$g \in \mathcal D_G - \{g^{slack}\}$}
\STATE $P_{g,0}^{(max)} \gets \mathbf{P}_g[t_0]$
\STATE $(P_{g,0}^{(dev)}, Q_{g,0}^{(dev)}) \gets \argmin_{(P, Q) \in \mathcal R_{g,0}} ||(P_{g,0}^{(max)}, q) - (P, Q)||$, with $q \sim U[\underline{Q}_g, \overline{Q}_g]$
\ENDFOR
\FOR{$d \in \mathcal D_{DES}$}
\STATE $SoC_{d,0} \sim U[\underline{SoC}_d, \overline{SoC}_d]$
\STATE ($P_{d,0}^{(dev)}, Q_{d,0}^{(dev)}) \gets (0, 0)$
\ENDFOR
\STATE $(P_{g^{slack},0}^{(dev)}, Q_{g^{slack},0}^{(dev)}) \gets$ solution of (\ref{eq:power_flow_equations}) with $V_0 = 1 \angle 0$
\end{algorithmic}
\end{algorithm}

\clearpage
\section{Experimental hyperparameters}
\label{app:hyperparameters}

The hyperparameters used for the experiments presented in Section \ref{sec:experiments} are summarized in Table \ref{tab:hyperparameters_ppo} for the PPO and in Table \ref{tab:hyperparameters_sac} for the SAC algorithms. Both implementations were taken from the Stable Baselines 3 library \cite{stable-baselines3}. The horizon $T$ is the maximum number of steps per episode used during training. 

\begin{table}[h]
\centering
 \begin{tabular}[]{ll} \toprule
 Hyperparameter & Value \\
 \midrule
 Horizon ($T$) & $5000$ \\
 Adam learning rate & $3\times 10^{-4}$\\
 Steps per update & 2048 \\
 Num. epochs & 10 \\
 Minibatch size & 64 \\
 GAE parameter ($\lambda$) & 0.95 \\
 Clipping parameter ($\epsilon$) & 0.2 \\
 VF coeff. c1 & 0.5 \\
 Entropy coeff. c2 & 0.0 \\
 Normalized observations & True \\
 \bottomrule
\end{tabular}
\vspace{0.2cm}
\caption{PPO hyperparameters.}
\label{tab:hyperparameters_ppo}
\end{table}

\begin{table}[h]
\centering
 \begin{tabular}[]{ll} \toprule
 Hyperparameter & Value \\
 \midrule
 Horizon ($T$) & $5000$ \\
 Adam learning rate & $3\times 10^{-4}$ \\
 Replay buffer size & $10^6$ \\
 Steps per update & 1 \\
 Minibatch size & 256 \\
 Target smoothing coefficient ($\tau$) & 0.005 \\
 Target update interval & 1 \\
 Gradient steps & 1 \\
 Entropy regularization coefficient & 'auto'\\
 Normalized observations & True \\
 \bottomrule
\end{tabular}
\vspace{0.2cm}
\caption{SAC hyperparameters.}
\label{tab:hyperparameters_sac}
\end{table}

\end{appendices}

\end{document}